\documentclass[10pt,journal,compsoc]{IEEEtran}

\usepackage{amsmath,amssymb,amsfonts}
\usepackage{mathrsfs}
\usepackage{algorithmic}
\usepackage{algorithm}
\usepackage{array}
\usepackage[caption=false,font=normalsize,labelfont=sf,textfont=sf]{subfig}
\usepackage{textcomp}
\usepackage{stfloats}
\usepackage{url}
\usepackage{verbatim}
\usepackage{graphicx}
\usepackage{cite}
\usepackage{color}

\usepackage{booktabs}
\usepackage{multirow}

\usepackage{balance}

\begin{document}
\title{On Jointly Optimizing Partial Offloading and SFC Mapping: A Cooperative Dual-agent Deep Reinforcement Learning Approach}

\author{Xinhan~Wang,
        Huanlai~Xing*,~\textit{Member}, IEEE,
        Fuhong~Song,
        Shouxi~Luo,~\textit{Member}, IEEE, 
        Penglin~Dai,~\textit{Member}, IEEE,
        and Bowen~Zhao % <-this % stops a space
\IEEEcompsocitemizethanks{\IEEEcompsocthanksitem X. Wang, H. Xing*, F. Song, S. Luo, P. Dai, and B. Zhao are with School of Computing and Artificial Intelligence, Southwest Jiaotong University, Chengdu 611756, China. (Corresponding author: Huanlai Xing) \protect\\
E-mail: xhwang@my.swjtu.edu.cn, hxx@home.swjtu.edu.cn, fhs@my.swjtu.edu.cn,  sxluo@swjtu.edu.cn, penglindai@swjtu.edu.cn, cn16bz@icloud.com.
%\protect\\
%*Corresponding author.
}% <-this % stops an unwanted space
\thanks{Manuscript received --; revised --.}}

\markboth{}%
{Wang \MakeLowercase{\textit{et al.}}: On Jointly Optimizing Partial Offloading and SFC Mapping: A Cooperative Dual-agent Deep Reinforcement Learning Approach}

\IEEEtitleabstractindextext{%
\begin{abstract}
Multi-access edge computing (MEC) and network function virtualization (NFV) are promising technologies to support emerging IoT applications, especially those computation-intensive. In NFV-enabled MEC environment, service function chain (SFC), i.e., a set of ordered virtual network functions (VNFs), can be mapped on MEC servers. Mobile devices (MDs) can offload computation-intensive applications, which can be represented by SFCs, fully or partially to MEC servers for remote execution. This paper studies the partial offloading and SFC mapping joint optimization (POSMJO) problem in an NFV-enabled MEC system, where an incoming task can be partitioned into two parts, one for local execution and the other for remote execution. The objective is to minimize the average cost in the long term which is a combination of execution delay, MD's energy consumption, and usage charge for edge computing. This problem consists of two closely related decision-making steps, namely task partition and VNF placement, which is highly complex and quite challenging. To address this, we propose a cooperative dual-agent deep reinforcement learning (CDADRL) algorithm, where we design a framework enabling interaction between two agents. Simulation results show that the proposed algorithm outperforms three combinations of deep reinforcement learning algorithms in terms of cumulative and average episodic rewards and it overweighs a number of baseline algorithms with respect to execution delay, energy consumption, and usage charge.
\end{abstract}

\begin{IEEEkeywords}
Deep reinforcement learning, multi-access edge computing, network function virtualization, partial offloading, SFC mapping.
\end{IEEEkeywords}}

\maketitle

\IEEEraisesectionheading{\section{Introduction}\label{Introduction}}
\IEEEPARstart{D}{ue} to the advent of fifth generation (5G) networks and Internet of Things (IoT) era, the number of mobile devices (MDs), including smartphones, wearables, tablets, sensors, etc., has increased rapidly \cite{goudarzi2020application,liang2021multi}. Meanwhile, many computation-intensive IoT applications are emerging, such as augmented/virtual reality (AR/VR), interactive online gaming and object recognition. Such applications are usually highly dependent on the availability of computing and storage resources. When executing these applications, MDs are strictly constrained by limited computing, communication, and battery capacities \cite{siriwardhana2021survey}. Multi-access edge computing (MEC) has been considered as a promising solution to the problem above as computation offloading allows MDs to offload some of computation-intensive applications to MEC servers for execution, reducing the computing pressure on MDs \cite{yang2020computation,song2022offloading}. 

In MEC, computation offloading can be classified into two categories, namely binary and partial offloading \cite{siriwardhana2021survey}. In binary offloading, a task is executed either locally or remotely, i.e., ``all or none". In contrast, partial offloading partitions a task into two parts. One part is run on the MD and the other part is offloaded to the edge infrastructure (EI) for remote execution. Both parts can be executed in parallel, which helps to reduce application latency and energy consumption, thus significantly improves the quality of experience (QoE) \cite{lin2020survey}. Therefore, this paper focuses on the partial offloading model.

On the other hand, network function virtualization (NFV) has been envisaged as one of the fundamental technologies for next-generation networks, e.g., the sixth generation (6G) network and space-air-ground integrated network (SAGIN). NFV decouples network functions from proprietary hardware and implements them as virtual network functions (VNFs). Faced with the explosion of computation-intensive and delay-sensitive applications, enabling NFV in MEC ensures users' QoE and reduces both capital expenditure (CAPEX) and operational expense (OPEX) from the point of view of network service providers (NSPs), which is an inevitable trend of MEC \cite{ma2020virtual}. The European Telecommunications Standards Institute (ETSI) introduced an MEC reference architecture in \cite{etsi_mobile_2018}, where MEC was deployed as part of an NFV environment. In this architecture, a mobile edge application is represented by a set of ordered VNFs called service function chain (SFC) \cite{nguyen2020joint}. 

In the literature, most studies on NFV-enabled MEC systems focused on SFC mapping in EI environment. They considered placing VNFs on MEC servers in order to optimize the end-to-end delay, energy consumption or other performance metrics. Recently, jointly optimizing computation offloading and SFC mapping has attracted increasingly more research interests. Most of them, however, consider binary offloading. With respect to partial offloading, a task is partitioned into two parts that can be processed in parallel by two identical SFCs. The decision making for partial offloding and that for the corresponding VNF placement are closely related \cite{kim2018control,khoramnejad2021distributed}. Provisioning parti al offloading service in NFV-enabled MEC network with a specific SFC requirement poses many challenges. \textit{For example, how to partition the task to achieve optimal offloading performance? How to place the required VNFs on proper MEC servers to meet the service demands? And how to jointly optimize the partial offloading and SFC mapping in order to find an acceptable trade-off among execution delay, energy consumption and usage charge?}

The partial offloading and SFC mapping joint optimization (POSMJO) consists of two closely coupled decision-making steps, namely task partition and VNF placement. The task partition step decides the proportion of data to be offloaded for remote execution, while the VNF placement step deploys the corresponding VNFs on appropriate MEC servers. Both of them are NP-hard yet require realtime solution. Deep reinforcement learning (DRL) algorithms, which combine reinforcement learning (RL) with deep neural network (DNN), appear as visible approaches to NP-hard like computation offloading \cite{song2022offloading} and SFC mapping \cite{liu2021sfc}. On the other hand, the first step is the premise of the second one, and the result of the second step directly affects the performance of the first one. Thus, the two steps are complementary and closely related. To obtain an acceptable solution to the POSMJO problem demands for appropriate collaboration between them. Traditional DRL methods can handle either environments with discrete actions, e.g., deep Q-network (DQN) \cite{hessel2018rainbow}, or those with continuous actions, e.g., deep deterministic policy gradient (DDPG) \cite{lillicrap_ddpg_2015}. Task partition and VNF placement are continous and discrete optimization problems, respectively. Using a single-agent DRL to tackle the two steps above simultaneously is almost impossible. This motivates us to design a cooperative dual-agent DRL (CDADRL) algorithm to address the POSMJO problem, where two agents collaborate on dealing with the two closely related steps.

This paper studies the partial offloading and SFC mapping joint optimization (POSMJO) problem in the NFV-enabled MEC environment. Unlike existing studies focusing on binary offloading, an incoming task is partitioned into two parts, with one run locally and the other offloaded to and executed at the EI. Both parts need to be processed by the same SFC. A DRL algorithm with two collaborative agents is presented to address the problem. The main contributions of this paper are as follows.

\begin{itemize}
	\item  We formulate the POSMJO problem in the context of NFV-enabled MEC, where the execution delay, energy consumption, and usage charge are three objectives to minimize, subject to a number of constraints such as computing and bandwidth capacities, and task demand. The problem consists of two closely related decision-making steps, namely task partition and VNF placement. 
	
	\item A cooperative dual-agent DRL (CDADRL) algorithm is devised to address the POSMJO problem, where a framework with two environments, namely, the MD and EI environments, is used to support the interaction between two agents. One agent interacts with the MD environment for task partition, while the other interacts with the EI environment for VNF placement.
	
	\item We conduct extensive simulations to evaluate the performance of the proposed algorithm. The results demonstrate that CDADRL outperforms three combinations of state-of-the-art DRL algorithms in terms of cumulative and average episodic reward. In addition, the proposed algorithm obtains lowest execution delay, energy consumption and usage charge compared with four baseline algorithms with various offloading schemes.
	
\end{itemize}

The rest of the paper is organized as follows. The related work is reviewed in Section~\ref{Related Work}. Section~\ref{System Model} describes the system model and problem formulation. Then, Section~\ref{CDADRL Algorithm} introduces in detail the proposed CDADRL algorithm for the POSMJO problem. Section~\ref{Simulation Results and Discussion} evaluates the performance of the proposed algorithm. Section~\ref{Conclusion} concludes the paper and presents future work.

\section{Related Work} \label{Related Work}
Over the past few years, there have been increased efforts to study NFV-enabled MEC. Generally speaking, these studies can be classified into two categories, namely VNF placement optimization and computation offloading and VNF placement joint optimization.

\subsection{VNF Placement Optimization}
In NFV-enabled MEC systems, mobile edge applications are implemented as SFCs in the EI environment, where each SFC is a set of ordered VNFs deployed on a substrate network \cite{etsi_mobile_2018}. Huang \textit{et al.} \cite{huang2020reliability} considered the reliability-aware VNF placement in MEC and formulated it as an integer linear programming (ILP), where an approximation algorithm was proposed to maximize the network throughput. Chantre \textit{et al.} \cite{chantre2020location} studied the location problem for the provisioning of protected slices in NFV-enabled MEC. They modeled it as a multi-criteria optimization problem and proposed a multi-objective evolutionary non-dominated sorting genetic algorithm. Subramanya \textit{et al.} \cite{subramanya2020machine} converted the SFC placement problem in an MEC-enabled 5G network into an ILP model, where machine learning methods were adopted for network traffic prediction. Zheng \textit{et al.} \cite{zheng2020toward} studied a hybrid SFC embedding problem in the context of MEC and addressed it by a heuristic algorithm based on hybrid SFC embedding auxiliary graph. Yang \textit{et al.} \cite{yang2021delay} investigated the VNF placement and routing problem in edge clouds and proposed a randomized rounding approximation algorithm to minimize the maximum link load. Xu \textit{et al.} \cite{xu2021nfv} studied the provisioning of IoT applications in NFV-enabled MEC, where the IoT application and VNFs of each service request were consolidated into a single location. They devised a heuristic algorithm to jointly place the IoT application and its corresponding VNFs. Ma \textit{et al.} \cite{ma2022mobility} considered mobility-aware and delay-sensitive service provisioning in MEC, where an approximation algorithm was devised to maximize the network utility and throughput. 

Apart from the methods above, DRL-based approaches exhibit a rising trend. Shu \textit{et al.} \cite{shu2020deploying} introduced the problem of deploying VNFs in MEC with time-varying geographical of users, where a DRL-based method was employed to estimate the underlying wireless features affected by user mobility and improve the resource utilization of MEC servers. Dalgkitsis \textit{et al.} \cite{dalgkitsis2020dynamic} emphasized the automatic VNF placement problem in 5G network, where DDPG was used to minimize latency for ultra-reliable low-latency communication (uRLLC) services. Liu \textit{et al.} \cite{liu2021dynamic} investigated SFC orchestration problem in NFV/MEC-enabled IoT networks, where a DRL method was devised for dynamic SFC orchestration. Later on, the authors improved their work by leveraging DDPG and asynchronous advantage actor-critic (A3C) that handled small- and large-scale networks, respectively \cite{liu2021sfc}. Xu \textit{et al.} \cite{xu2022cloud} focused on SFC mapping for industrial IoT (IIoT), where a DQN-based online SFC deployment method was designed to balance the quality of IIoT services and resource consumption. Abouaomar \textit{et al.} \cite{abouaomar2021mean} investigated the resource provisioning problem for VNF placement and chaining in MEC. The authors proposed a mean-field game framework to model the behaviour of placing and chaining VNFs, and they leveraged an actor-critic DRL to learn the optimal placement and chaining policies.

\subsection{Computation Offloading and VNF Placement Joint Optimization}
Computation offloading is an effective way for 5G/6G networks to deal with computation-intensive and delay-sensitive tasks. In partial offloading, for an arbitrary task, the part for local execution and that for remote execution need to be processed by two identical SFCs, respectively. Therefore, the decision making process of computation offloading and that of VNF placement are closely related, forming a joint optimization problem. This problem has attracted increasingly more research attention from both academia and industry. For instance, Ma \textit{et al.} \cite{ma2020virtual} jointly optimized VNF placement and user requests assignment in MEC. If the computing capacity on each cloudlet was sufficient, an ILP solution was proposed to maximize the requests admission number; otherwise, two heuristic algorithms were developed to maximize the network throughput. Nguyen \textit{et al.} \cite{nguyen2020joint} investigated the joint optimization problem of computation offloading, SFC placement and resource allocation in NFV-enabled MEC, where a decomposition approach was employed to minimize the weighted normalized energy consumption and computing cost. Kim \textit{et al.} \cite{kim2018control} designed a dual-resource NFV-enabled MEC system, with the multi-path and sending rate control problem studied. A convex optimization method based on Lagrange dual theory was developed to handle the problem constraints, and an extragradient-based algorithm was proposed to control and split the service sending rate. Jia \textit{et al.} \cite{jia2017qos} formulated a task offloading and VNF placement problem in a metropolitan area network, where each offloaded task requested a specific VNF. The authors presented an online heuristic algorithm to maximize the requests admission number. Xu \textit{et al.} \cite{xu2019task} investigated the task offloading problem in MEC with quality of service (QoS) requirement and devised an online algorithm with competitive ratio. Jin \textit{et al.} \cite{jin2019computation} emphasized computation offloading of online games in a fog-enabled heterogeneous radio access network, where the games' interactive and uncertain features were modeled as probabilistic SFCs. They formulated the computation offloading problem as an ILP and proposed a heuristic algorithm to address it. Khoramnejad \textit{et al.} \cite{khoramnejad2021distributed} looked into the partial offloading problem in NFV-enabled MEC, where each VNF in the required SFC can be executed locally or offloaded to an MEC server. The authors adapted double DQN (DDQN) to the problem and obtained decent performance.

In this paper, we consider POSMJO in the context of NFV-enabled MEC. We distinguish our work from the existing ones as follows. 
\begin{itemize}
	\item  In terms of computation offloading model, most existing works focus on binary offloading \cite{jia2017qos,kim2018control,xu2019task,jin2019computation,ma2020virtual,nguyen2020joint}. However, some applications contain some non-offloadable part(s), e.g., user input, camera, and positioning request, that need to be executed on the MD \cite{mach2017mobile}. Therefore, binary offloading cannot be applied to these applications. There are a few studies on partial offloading, where one or more VNFs in an SFC are offloaded for remote execution \cite{khoramnejad2021distributed}. Nevertheless, none of the studies considered partitioning the incoming task into two parts, one for local execution and the other for remote execution. With the advent of 5G networks and imminent of 6G networks, it is inevitable to provision partial offloading in NFV-enabled MEC, making full use of the computation offloading advantages. Our work assumes that a task can be partitioned into two parts that go through two identical SFCs, separately. One part is executed on the local MD and the other at the edge in parallel.
	
	\item Most existing computation offloading and SFC mapping methods are based on heuristic \cite{zheng2020toward,xu2021nfv,jia2017qos,xu2019task,jin2019computation,ma2020virtual,nguyen2020joint}, approximation  \cite{huang2020reliability,yang2021delay,ma2022mobility}, convex optimization \cite{kim2018control}, and meta-heuristic algorithms \cite{chantre2020location}. However, these methods are usually highly complex and time-consuming, which are not suitable for dynamic MEC scenarios. On the other hand, DRL can predict some environmental information through offline model training and make near-optimal decisions. It is more suitable for dynamic and complex environments. Therefore, it is natural to adopt DRL to cope with the POSMJO problem. Methods based on DRL \cite{shu2020deploying,dalgkitsis2020dynamic,liu2021dynamic,liu2021sfc,xu2022cloud,khoramnejad2021distributed} demonstrate their ability for efficiently handling various decision making processes in time-varying MEC systems. This paper considers the relation between the two closely related decision-making steps, namely task partition and VNF placement, and it devises a DRL algorithm with two cooperative agents to handle the POSMJO problem.
	
\end{itemize}

\begin{table}[!t]
	\caption{Summary of Main Notations.\label{tab:notations}}
	\centering
	\begin{tabular}{ll}
		\hline
		Notation & Definition\\
		\hline
		\multicolumn{2}{l}{Notation used in system model.} \\
		\hline
		$br_i^t$ & Required bandwidth capacity between $f_i^t$ and $f_{i+1}^t$ \\
		$bw_e^t$ & Available bandwidth capacity of $ e $  at $t$\\
		$c^t$ & Number of CPU cycles required to execute $u^t$\\
		$cp(f_i^t)$ & Required computing capacity for execution $f_i^t$\\
		$cp_{MD}^t$ & Available computing capacity of the MD at $t$\\
		$cp_v^t$ & Available computing capacity of BS $ v $  at $t$\\
		$d^t$ & Input data size of $u^t$\\
		$D_{max}^t$ & Execution deadline of $u^t$\\
		$DC^t$ & Execution delay of $u^t$\\
		$DE^t$ & Edge computing delay\\
		$DI(F_k^t)$ & Instantiation Delay of $F_k^t$\\
		$DL^t$ & Local computing delay\\
		$DT^t$ & Transmission delay of $u^t$\\
		$down^t$ & Achievable downlink transmission rate\\
		$EC^t$ & Energy consumption of the MD for executing $u^t$\\
		$EE^t$ & Energy consumption for edge computing\\
		$EL^t$ & Energy consumption for local computing\\
		%$EU^t$ & Energy consumption for transmitting data from the MD to $v_{MD}^t$\\ 
		\multirow{2}{*}{$EU^t$} & Energy consumption for transmitting data from \\
		& the MD to $v_{MD}^t$\\ 
		$F^t$ & Required SFC corresponding to $u^t$\\
		$F_k^t$ & The $k$-th group of $F^t$\\
		$f_i^t$ & The $i$-th VNF in $F^t$\\
		$f_{k,j}^t$ & The $j$-th VNF in $F_k^t$\\
		$G=(V,E)$ & Edge infrastructure\\
		$g^t_{down}$ & Downlink channel gain between $v_{MD}^t$ and the MD\\
		$g^t_{up}$ & Uplink channel gain between the MD and $v_{MD}^t$\\
		$h_i^t$ & The amount of output data after $u^t$ is processed by $f_i^t$\\
		$v_{MD}^t$ & The closest BS to the MD\\
		$N^t$ & Number of VNFs in $F^t$ \\
		$N_k^t$ & Number of VNFs in $F_k^t$ \\
		$p_{MD}^{tr}$ & Transmission power of the MD\\
		$p_{MD}^{re}$ & Receiving power of the MD\\
		$p_{v_{MD}^t}^{tr}$ & Transmission power of $v_{MD}^t$\\
		$\rho_i^t$ & BS selected to host VNF $f_i^t \in F^t$\\
		$UC^t$ & Usage charge for edge computing\\
		$u^t$ & Incoming task of the MD at $t$\\
		$up^t$ & Achievable uplink transmission rate\\
		$W$ & Wireless bandwidth\\
		$x^t$ & Offloading ratio\\
		$\xi_i$ & Ratio of $h_i^t$ to $d^t$\\
		$\eta^2$ & Noise power\\
		\hline
		\multicolumn{2}{l}{Notation used in reinforcement learning.} \\
		\hline
		$a$ & Action\\
		$\mathcal{A}$ & Action space\\
		$\mathbf{A_o}$ & Agent in charge of task partition\\
		$\mathbf{A_p}$ & Agent in charge of VNF placement\\
		$\mathcal{B}$ & Replay buffer\\
		$r(s,a)$ & Reward function\\
		$s$ & State\\
		$\mathcal{S}$ & State space\\
		$\gamma$ & Discount factor\\
		\hline
	\end{tabular}
\end{table}

\section{System Model and Problem Formulation} \label{System Model}

We consider an NFV-enabled MEC system consisting of one MD and the edge infrastructure (EI). The EI is represented by an undirected graph $ G=(V,E) $, where $V$ is the set of base stations (BSs) and $E$ is the set of wired links between BSs. Each $ v \in V $ is connected to an MEC server with limited computing capacity. Each MEC server can host multiple VNFs to implement the required mobile edge applications and provide computation offloading services for the MD. We consider a discrete-time system where $ t \in \lbrace 1,...,T \rbrace $ is the $t$-th time slot and $T$ is the monitoring period in terms of numbers of time slots. Let $ cp_v^t $ represent the available computing capacity of BS $v\in V$ in time slot $t$. Denote the available bandwidth capacity of link $ e \in E $ in time slot $t$ by $ bw_e^t $. In the system, the BS connected with the most powerful MEC server is designated as the central management node that runs the NFV management and orchestration (NFV-MANO) framework \cite{etsi_mobile_2018}. The NFV-MANO is responsible for VNF orchestration and life cycle management, including VNF placement, routing, and so on. 

The MD communicates with the EI through the closest BS $v_{MD}^t$ to receive computation offloading service \cite{shakarami2020survey}. In partial offloading, the input data of a task can be arbitrarily partitioned into two parts for local and edge computing \cite{lin2020survey,ren2017partial,li2020energy}. The two parts are processed by two identical SFCs, one deployed on the MD and the other at the edge \cite{nguyen2020joint}. Let $ u^t = \lbrace F^t, d^t, c^t, D_{max}^t \rbrace$ be the task of the MD in time slot $t$, where $ F^t $ represents the required SFC, $d^t$ is the input data size, $c^t$ is the number of CPU cycles required to execute $u^t$, and $D_{max}^t$ is the execution deadline of $u^t$. Let $f_i^t$, $i=1,...,N^t$, be the $i$-th VNF in $F^t$ and the computing capacity consumed for implementing $f_i^t$ is $ cp(f_i^t) $, where $N^t$ is the number of VNFs in $F^t$. Assume the delay for instantiating $f_i^t$ on the MD or an MEC server is $DI(f_i^t)$. The required bandwidth capacity between $f_i^t$ and $f_{i+1}^t$ is denoted by $br_i^t$. The data flow of $u^t$ needs to be processed by $ f_1^t,...,f_{N^t}^t$. Let $x^t \in [0,1]$ denote the offloading ratio for partitioning $u^t$, indicating how much data in $d^t$ is offloaded to the edge, i.e., $x^t \cdot d^t$ is the amount of data to be executed remotely. Let $h_i^t$, $i=1,...,N^t$, be the amount of output data after $u^t$ is processed by $f_i^t$. Let $ \xi_i^t $ be the ratio of $h_i^t$ to $d^t$. $\xi_i^t \cdot d^t$ is the amount of output data after $u^t$ is processed by $f_i^t$.

\begin{figure*}[!t]
	\centering
	\includegraphics[width=7in]{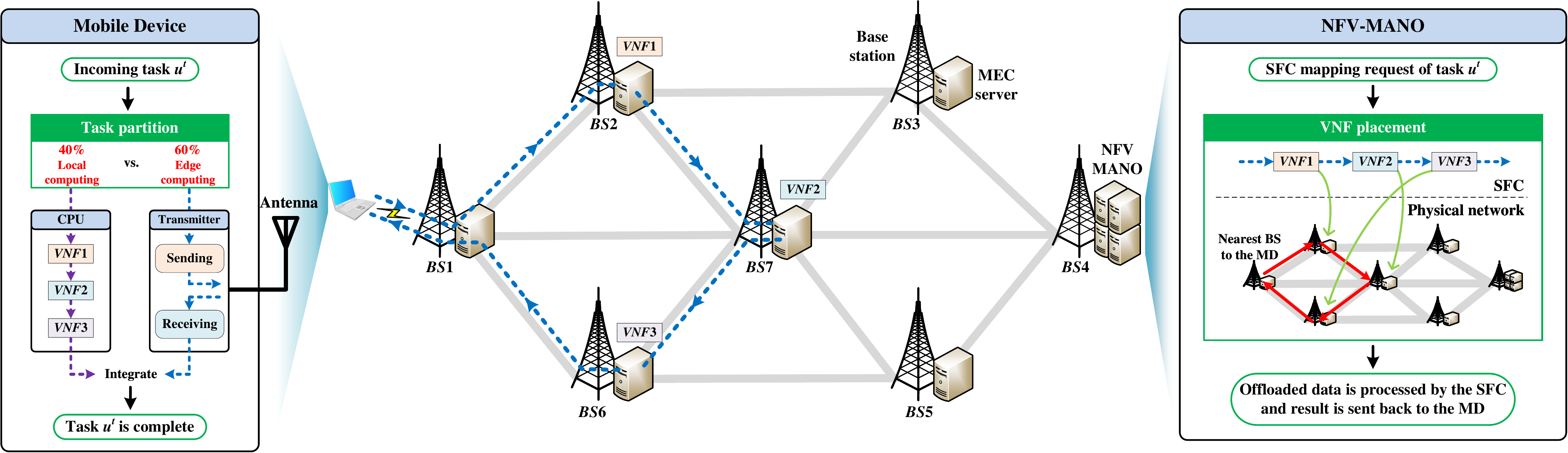}
	\caption{Example of joint partial offloading and SFC mapping in the NFV-enabled MEC scenario.}
	\label{fig-MEC_system}
\end{figure*}

Fig.~\ref{fig-MEC_system} depicts an example of joint partial offloading and SFC mapping in the NFV-enabled MEC scenario. There are seven BSs in the EI and the NFV-MANO framework is running on $BS4$. There is an MD connecting to $BS1$ and an incoming task $u^t$ arrives in time slot $t$. The input data of task $u^t$ goes through $VNF1\rightarrow VNF2\rightarrow VNF3$ to implement some mobile edge application. Assume the MD partitions the input of $u^t$ into two parts with $ x^t=0.6 $, i.e., 40\% of the input task data is processed on the MD while the rest is offloaded to the EI for remote execution. The NFV-MANO selects $BS2$, $BS6$, and $BS7$ to host these VNFs, and at the meantime the MD instantiates the required VNFs locally subject to the availability of its computing resources. The offloaded data is processed at the EI and the result is delivered back to the MD. The MD integrates the two parts and then task $u^t$ is complete.

According to the Shannon–Hartley theorem \cite{tse2005fundamentals}, the achievable uplink transmission rate from the MD to BS $v_{MD}^t$ is
\begin{equation}
	\label{eq-up_rate}
	up^t = W \cdot \log_{2}{\left( 1+ \frac{p_{MD}^{tr} \cdot g^t_{up}}{\eta^2}  \right)} ,
\end{equation}
where $W$, $p_{MD}^{tr}$, and $\eta^2$ are the wireless bandwidth, transmission power of the MD, and noise power, respectively. $g^t_{up}$ is the uplink channel gain between the MD and $v_{MD}^t$. Similarly, the achievable downlink transmission rate is
\begin{equation}
	\label{down_rate}
	down^t = W \cdot \log_{2}{\left( 1+ \frac{p_{v_{MD}^t}^{tr} \cdot g^t_{down}}{\eta^2}  \right)} ,
\end{equation}
where $p_{v_{MD}^t}^{tr}$ is the transmission power of BS $v_{MD}^t$, and $g^t_{down}$ is the downlink channel gain between $v_{MD}^t$ and the MD. The main notations used in this paper are summarized in Table~\ref{tab:notations}.

\subsection{Local Computing}
Note that we  may not instantiate all VNFs in $F^t$ simultaneously on the MD, due to the limitation of its computing capacity. We divide $F^t$ into $\psi^t$ groups according to the MD's computing capacity $cp_{MD}^t$ in time slot $t$. To make full use of the MD's computing capacity, the MD releases its corresponding computing resources occupied immediately after the input data has been processed by all VNFs in the $k$-th group $F_k^t$ orderly. The relation between $F^t$ and $F_k^t$ is defined as
\begin{equation}
	\label{eq-group_unify}
	\bigcup_{k=1}^{\psi^t} {F_k^t} = F^t.
\end{equation}
In time slot $t$, the computing capacity consumed by $F_k^t$ cannot exceed the MD's computing capacity $cp_{MD}^t$, i.e.,
\begin{equation}
	\label{eq-group_capacity}
	\sum_{j=1}^{N_k^t} {cp(f_{k,j}^t)} \leq cp_{MD}^t ,
\end{equation}
where $f_{k,j}^t \in F_k^t$ is the $j$-th VNF in $F_k^t$, and $N_k^t$ is the number of VNFs in $F_k^t$. Take Fig.~\ref{fig-MEC_system} as an example. If the MD's computing capacity can only afford to host $VNF1$ and $VNF2$ at the same time, we instantiate them at first. After the task data is processed by them, the MD releases the computing resources occupied by $VNF1$ and $VNF2$. And then it instantiates $VNF3$ to complete the local execution of $u^t$.

After all VNFs in $F_k^t$ are instantiated, the amount of input data of $u^t$, $(1-x^t) \cdot d^t$, is processed by every VNF in $F_k^t$ in the right order. That $f_{k,j}^t$ processes the task data incurs a processing delay, $DP_{k,j}^t$, written as
\begin{equation}
	\label{eq-group_process_delay_ij}
	\begin{split}
		DP_{k,j}^t &= \begin{cases} 
			\frac{\chi}{cp(f_{k,j}^t)}, & \text {if $k=1$, $j=1$} 
			\\ \frac{\xi_{k-1,N_{k-1}^t}^t \cdot \chi}{cp(f_{k,j}^t)}, & \text{if $k=2,...,\psi^t$, $j=1$}
			\\ \frac{\xi_{k,j-1}^t \cdot \chi}{cp(f_{k,j}^t)}, & \text{if $k=1,...,\psi^t$, $j=2,...,N_k^t$}
		\end{cases}
	\end{split}	
\end{equation}
where $\chi = (1-x^t) \cdot d^t \cdot c^t$, and $\xi_{k,j}^t$ is the ratio of $h_{k,j}^t$ to $d^t$, and $h_{k,j}^t$ is the amount of output data after $u^t$ is processed by $f_{k,j}^t$. Let $DI(F_k^t)$ be the instantiation delay of $F_k^t$, where $DI(F_k^t)$ is equal to the maximum VNF instantiation delay in $F_k^t$, i.e., $DI(F_k^t) = \max {\lbrace DI(f_{k,j}^t) \mid \forall f_{k,j}^t \in F_k^t \rbrace}$. Based on the instantiation delay and processing delay of each group $F_k^t$, we calculate the local computing delay, $DL^t$, by
\begin{equation}
	\label{eq-local_delay}
	DL^t = \sum_{k=1}^{\psi^t} {\left( DI(F_k^t) + \sum_{j=1}^{N_k^t} {DP_{k,j}^t} \right)} .
\end{equation}

The energy consumption for local computing, $EL^t$, is defined as
\begin{equation}
	\label{eq-local_energy}
	EL^t = \sum_{i=1}^{N^t} {\xi_{i-1}^t \cdot(1-x^t) \cdot d^t \cdot c^t \cdot \kappa \cdot (cp(f_i^t))^2} ,
\end{equation}
where $\kappa$ is the effective capacitance coefficient dependent on the chip architecture used and $\xi_0^t = 1$.

\subsection{Edge Computing}
In time slot $t$, a proportion of the input data of $u^t$, $x^t \cdot d^t$, needs to be offloaded to the EI for remote execution. At first, NFV-MANO generates an SFC mapping scheme for $F^t$ according to the MD's offloading decision and issues relevant instructions to the corresponding BSs. Then, these BSs instantiate the requested VNFs. Denote by $\rho_i^t$ the BS selected to host VNF $f_i^t \in F^t$. The instantiation delay of $F^t$ at the EI, $DI(F^t)$, is defined as $DI(F^t) = \max {\lbrace DI(f_i^t) \mid \forall f_i^t \in F^t \rbrace}$. Meanwhile, considering that any two BSs in the system are close to each other, this paper ignores the propagation delay. 

The transmission delay of task $u^t$ consists of three parts: (1) the data transmitted through the wireless uplink channel from the MD to BS $v_{MD}^t$, (2) the data transmitted through wired links between BSs, and (3) the data transmitted through the wireless downlink channel from BS $v_{MD}^t$ to the MD. Thus, the total transmission delay of task $u^t$ is defined as
\begin{equation}
	\label{eq-edge_tr_delay}
	\begin{split}
		DT^t = 
		&\frac{x^t \cdot d^t}{up^t} + 
		\sum_{e\in E_1^t} {\frac{x^t \cdot d^t}{bw_e^t}} +
		\sum_{i=1}^{N^t-1} {\frac{\xi_i^t \cdot x^t \cdot d^t}{br_i^t}} + \\
		&\sum_{e\in E_N^t} {\frac{\xi_{N^t}^t \cdot x^t \cdot d^t}{bw_e^t}} + 
		\frac{\xi_{N^t}^t \cdot x^t \cdot d^t}{down^t}
	\end{split} ,
\end{equation}
where $E_1^t$ is the link set of the path from $v_{MD}^t$ to the BS hosting $f_1^t$, and $E_N^t$ is the link set of the path from the BS hosting $f_{N^t}^t$ to BS $v_{MD}^t$. Note that the transmission path between BSs involved in this paper is generated by the shortest path method, e.g., Dijkstra's algorithm \cite{dijkstra_note_1959}.

The total processing delay incurred by VNFs $ f_1^t,...,f_{N^t}^t$ processing the offloaded data at the EI, $DP_{E}^t$, is defined by
\begin{equation}
	\label{eq-edge_pr_delay}
	DP_{E}^t = \sum_{i=1}^{N^t} {\frac{\xi_{i-1}^t \cdot x^t \cdot d^t \cdot c^t} {cp(f_i^t)}}.
\end{equation}

Based on the instantiation, transmission and processing delays, $DI(F^t)$, $DT^t$ and $DP_{E}^t$, we calculate the edge computing delay, $DE^t$, by
\begin{equation}
	\label{eq-edge_delay}
	DE^t = DI(F^t) + DT^t + DP_{E}^t.
\end{equation}

The energy consumption for transmitting data from the MD to $v_{MD}^t$, $EU^t$, is defined as
\begin{equation}
	\label{eq-up_energy}
	EU^t = \frac{x^t \cdot d^t}{up^t} \cdot p_{MD}^{tr}.
\end{equation}

Let $ED^t$ be the energy consumption for the MD to receive the result data from $v_{MD}^t$ when the offloaded data is completely executed. $ED^t$ is calculated by
\begin{equation}
	\label{eq-down_energy}
	ED^t = \frac{\xi_{N^t}^t \cdot x^t \cdot d^t}{down^t} \cdot p_{MD}^{re} ,
\end{equation}
where $p_{MD}^{re}$ is the receiving power of the MD. The energy consumption for edge computing,  $EE^t$, is defined as
\begin{equation}
	\label{eq-edge_energy}
	EE^t = EU^t + ED^t .
\end{equation}

In addition, this paper considers usage charge when edge computing incurs. If a BS hosts a VNF $f_i^t \in F^t$, the system operator charges the MD because it rents the computing resources at the EI. The usage charge is calculated based on the price per time unit at the computing capacity $cp(f_i^t)$, which is defined as $\Delta(f_i^t) = e^{-\alpha} \cdot (e^{cp(f_i^t)} -1) \cdot \beta $, where $\alpha$ and $\beta$ are two coefficients  \cite{nguyen2020joint}. The usage charge for hosting $F^t$ at the EI is calculated by
\begin{equation}
	\label{eq-edge_charge}
	UC^t = \sum_{i=1}^{N^t} {\frac{\xi_{i-1}^t \cdot x^t \cdot d^t \cdot c^t} {cp(f_i^t)}} \cdot \Delta(f_i^t) .
\end{equation}

\subsection{Problem Formulation}
The POSMJO problem consists of two closely related decision-making steps, namely task partition and VNF placement. The first step decides the amount of task data to be offloaded to the EI. The second one places the corresponding VNFs on BSs to process the offloaded data. The first step is the premise of the second one, and the result of the second one directly affects the performance of the first one. Thus, the two steps are complementary and closely related. 

This paper considers three important indicators for measuring the offloading performance, i.e., task execution delay, energy consumption of the MD, and usage charge for edge computing. The execution delay directly determines the task completion time. Energy consumption is important as the MD is with limited battery. Usage charge depends on the budget of the MD for computation offloading service. Hence, the POSMJO problem is multi-objective in nature, with each performance indicator to be minimized. This paper adopts the weighted sum method \cite{cho_survey_2017,zajac_objectives_2021}, to aggregate the three performance indicators into a single-objective cost function to minimize, which is defined as 

\begin{equation}
	\label{eq-cost}
	Cost(u^t) = \omega_1 DC^t + \omega_2 EC^t + \omega_3 UC^t ,
\end{equation}
where $\omega_1 + \omega_2 + \omega_3 = 1$, and $DC^t$ is the execution delay of task $u^t$, defined as
\begin{equation}
	\label{eq-delay}
	DC^t = \max{\lbrace DL^t,DE^t \rbrace}.
\end{equation}
$DC^t$ is equal to $DL^t$ if the local computing delay $DL^t$ is larger than edge computing delay $DE^t$, and $DC^t$ equals $DE^t$, otherwise. $EC^t$ is the energy consumption of the MD executing task $u^t$, which is the summation of the energy consumption incurred by local and edge computing, that is, 
\begin{equation}
	\label{eq-energy}
	EC^t = EL^t + EE^t.
\end{equation}

The POSMJO problem aims to minimize the average long-term cost, written as,
\begin{equation}
	\label{eq-objective}
	\underset{x^t, \rho_i^t}{\text{min}} \frac{1}{T} \sum_{t=1}^{T} Cost(u^t) ,
\end{equation}
subject to:
\begin{align}
	\label{eq-constraint}
	&C1: x^t \in [0,1] , \nonumber \\
	&C2: \rho_i^t \in V, i=1,...,N^t , \nonumber \\
	&C3: DL^t \leq D_{max}^t , \nonumber \\
	&C4: cp(f_i^t) \leq cp_{MD}^t, i=1,...,N^t , \nonumber \\
	&C5: cp(f_i^t) \leq cp_{\rho_i^t}^t, i=1,...,N^t , \nonumber \\
	&C6: br_i^t \leq \min{\lbrace bw_e^t \mid \forall e \in E_{i,i+1}^t \rbrace} , i=1,...,N^t-1 , \nonumber \\
	&C7: DE^t \leq D_{max}^t . \nonumber
\end{align}
Constraint $C1$ specifies that an arbitrary amount of the input data of $u^t$ can be offloaded to the EI and executed there. Constraint $C2$ specifies that an arbitrary VNF $f_i^t \in F^t$ can be instantiated on any BS. Constraints $C3$ and $C7$ mean that the local computing delay and edge computing delay are no larger than the execution deadline of $u^t$. Constraints $C4$ and $C5$ ensure that the computing resouces consumed by $f_i^t \in F^t$ cannot exceed the available computing capacity of the MD and the selected BS, respectively. Constraint $C6$ guarantees that for any link $e \in E$, the total bandwidth consumption caused by a data-flow cannot exceed its available bandwidth capacity, where $E_{i,i+1}^t$ is the link set of the path between the BSs hosting $f_i^t$ and $f_{i+1}^t$.

\section{Cooperative Dual-agent DRL Algorithm} \label{CDADRL Algorithm}

\begin{figure*}[!t]
	\centering
	\includegraphics[width=7in]{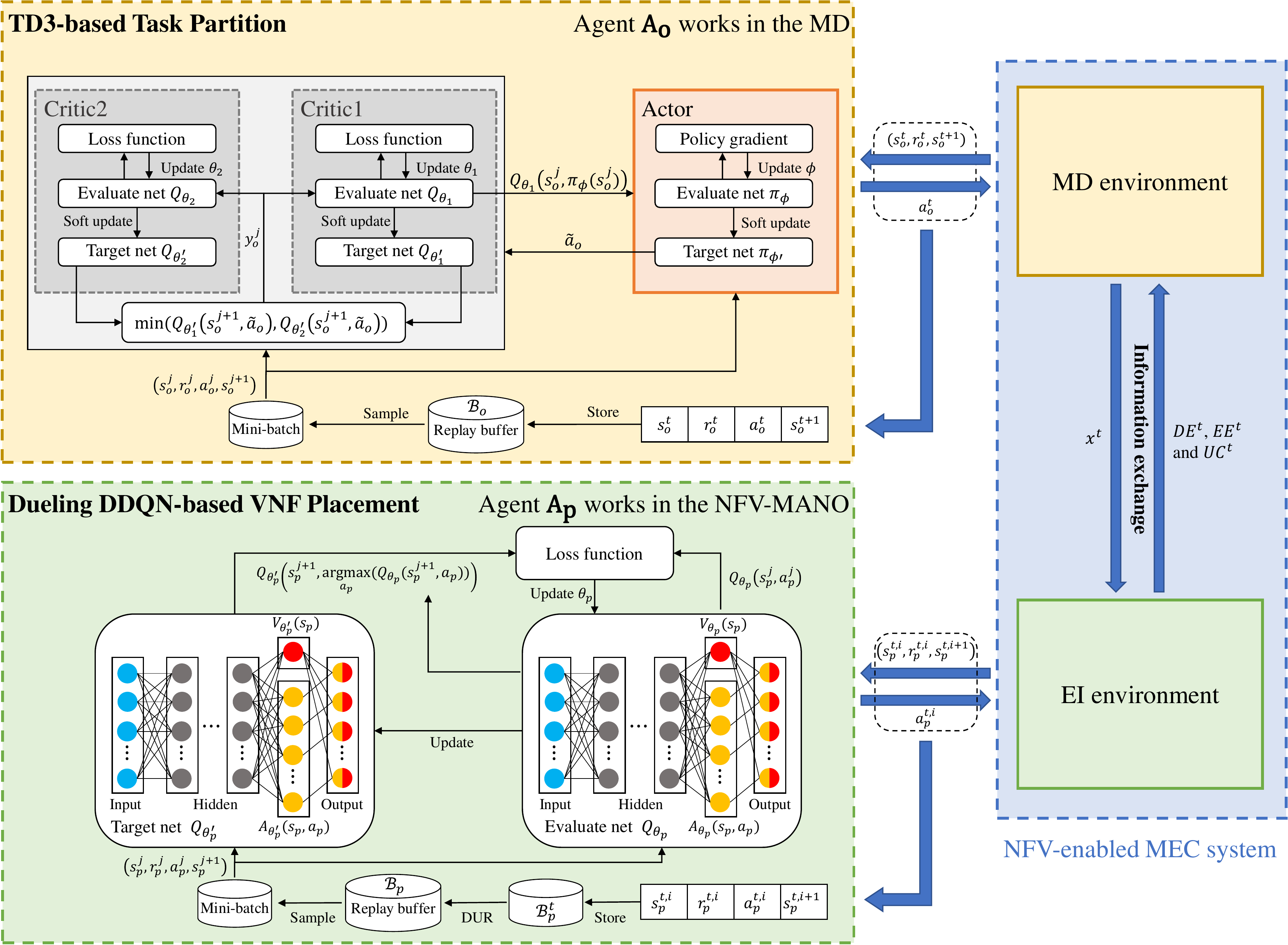}
	\caption{Framework of the proposed CDADRL algorithm.}
	\label{fig-framework}
\end{figure*}

We devise a cooperative dual-agent DRL (CDADRL) algorithm to realize the two closely related decision-making steps. The framework of the proposed CDADRL algorithm is depicted in Fig.~\ref{fig-framework}. In CDADRL, the NFV-enabled MEC system consists of two environments, namely the MD and EI environments, respectively. Let $\mathbf{A_o}$ and $\mathbf{A_p}$ denote the two agents interacting with environments MD and EI, respectively. Based on twin delayed DDPG (TD3), agent $\mathbf{A_o}$ is in charge of task partition, and it works in the MD. Based on dueling DDQN,   $\mathbf{A_p}$ is in charge of VNF placement, and it works in the NFV-MANO node. Agents $\mathbf{A_o}$ and $\mathbf{A_p}$ can gain their interest information through information exchange process between the MD and EI environments.

\subsection{MDP Formulation}
A DRL-solvable problem is usually modeled as a Markov decision process (MDP). An MDP can be denoted by a 3-tuple $ \left\langle \mathcal{S}, \mathcal{A}, r \right\rangle $, where $\mathcal{S}$ is the state space, and $\mathcal{A}$ is the action space, and $r(s,a)$ is the reward function, $s\in \mathcal{S}, a\in \mathcal{A}$. In time slot $t$, an agent at state $s^t$ takes an action $a^t$ based on a policy $\varphi$, where $\varphi$ is a specific state-to-action mapping, $\varphi:\mathcal{S} \rightarrow \mathcal{A}$. Then, the environment transits to a new state $s^{t+1}$ and returns an immediate reward $r^t \leftarrow r(s^t, a^t)$. The agent aims at learning a policy that maximizes future rewards, through interacting with the environment. The state-action value function corresponding to policy $\varphi$, $Q_\varphi(s,a)$, can be described as
\begin{equation}
	\label{eq-Q}
	Q_\varphi(s,a)	 = \mathbb{E}_\varphi \left[ \left. \sum_{k=t}^{T} \gamma^{(k-t)} \cdot r^k \right| s^t=s, a^t=a \right] ,
\end{equation}
where $\gamma\in [0,1]$ is a discount factor reflecting how important the future rewards are.

\subsubsection{MDP for Mobile Device Environment}
Given the MD environment, we train $\mathbf{A_o}$ to find an optimal policy on task partition that minimizes the execution delay, energy consumption and usage charge. The state space, action space, and reward space associated with the MD environment are defined as follows.

1) State space:
\begin{equation}
	\label{eq-state_o}
	\mathcal{S}_o = \lbrace \left. s_o^t \right| s_o^t=(F^t, d^t, c^t, D_{max}^t, cp_{MD}^t), t=1,...,T \rbrace , 
\end{equation}
where $s_o^t$ is the state of the MD environment in time slot $t$. $s_o^t$ is a vector of the required SFC of $u^t$, input data size of $u^t$, number of CPU cycles required to execute $u^t$, execution deadline of $u^t$, and available computing capacity of the MD.

2) Action space:
\begin{equation}
	\label{eq-action_o}
	\mathcal{A}_o = \lbrace \left. a_o^t \right| a_o^t \in [0,1], t=1,...,T \rbrace , 
\end{equation}
where $a_o^t$ is an action that $\mathbf{A_o}$ takes in $s_o^t$. In time slot $t$, the MD partitions the input data of task $u^t$ into local and edge computing parts. We set $x^t=a_o^t$. If $x^t=0$, task $u^t$ is executed on the MD locally. Otherwise, a proportion of the input data of $u^t$, $x^t \cdot d^t$, is executed at the EI, while the rest of the input data is processed on the MD.

3) Reward function:
Our objective is to minimize the average long-term cost consisting of the execution delay $DC^t$, energy consumption $EC^t$, and usage charge $UC^t$. So the reward function of the MD environment is set to the negative value of $Cost(u^t)$, defined as
\begin{equation}
	\label{eq-r_o}
	r_o^t = \begin{cases}
		-Cost(u^t), & \text {if constraints $C1-C7$ are met} \\
		-\varrho, & \text{otherwise}
	\end{cases} ,
\end{equation}
where $Cost(u^t)=\omega_1 DC^t + \omega_2 EC^t + \omega_3 UC^t$, and $\varrho$ is a sufficiently large number. Note that, $EC^t$ and $UC^t$ are obtained after the results are received from the EI environment. $-\varrho$ in reward function~\eqref{eq-r_o} is used to tell $\mathbf{A_o}$ the importance of choosing actions that satisfy constraints $C1-C7$.

\subsubsection{MDP for Edge Infrastructure Environment}
Given the EI environment, we train $\mathbf{A_p}$ to find an optimal policy on VNF placement that minimizes the edge computing delay and usage charge. As aforementioned, a solution to VNF placement specifies where each VNF in a given SFC is placed. Within time slot $t$, VNF placement of $u^t$ can be divided into $N^t$ stages. At stage $i$, $i=1,...,N^t$, $\mathbf{A_p}$ only decides where to place the $i$-th VNF in $F^t$ at the EI. The state space, action space, and reward space associated with the EI environment are defined as follows.

1) State space:
\begin{equation}
	\label{eq-state_p}
	\begin{split}
		\mathcal{S}_p = \lbrace \left. s_p^{t,i} \right| s_p^{t,i}=&(F^t, d^t, D_{max}^t, v_{MD}^t, x^t,  L^{t,i},	\\ &cp_{V}^{t,i}, bw_{E}^{t,i} ), 	t=1,...,T \rbrace  ,
	\end{split}
\end{equation}
where $s_p^{t,i}$ is the state of the EI environment at stage $i$ of $t$. In state $s_p^{t,i}$, $F^t$, $d^t$, and $D_{max}^t$ are the required SFC, input data size, and execution deadline of $u^t$, respectively. $v_{MD}^t$ is the closest BS to the MD, and $x^t$ is the offloading ratio. $L^{t,i}= (\rho_1^t,...,\rho_{N^t}^t )$ is used to identify the BSs selected to host VNFs $f_i^t \in F^t$. Because $\rho_i^t$ cannot be obtained when $\mathbf{A_p}$ observes $s_p^{t,i}$, we set $\rho_i^t=-1$. $cp_{V}^{t,i} = \lbrace cp_v^{t,i} \mid \forall v \in V \rbrace $ represents the available computing capacities of all BSs. $bw_{E}^{t,i} = \lbrace bw_e^{t,i} \mid \forall e \in E \rbrace $ stands for the available bandwidth capacities of all wired links between BSs.

2) Action space:
\begin{equation}
	\label{eq-action_p}
	\mathcal{A}_p = \lbrace \left. a_p^{t,i} \right| a_p^{t,i} \in V, t=1,...,T \rbrace , 
\end{equation}
where $a_p^{t,i}$ is an action $\mathbf{A_p}$ takes in $s_p^{t,i}$. At each stage, we set $\rho_i^t = a_p^{t,i}$, indicating BS $\rho_i^t$ has been selected to host VNF $f_i^t$. Therefore, we can place all VNFs $f_i^t \in F^t$ after $N^t$ stages.

3) Reward function:
Agent $\mathbf{A_p}$ is trained to minimize the edge computing delay $DE^t$ and usage charge $UC^t$ in time slot $t$, with a number of constraints corresponding to the VNF placement step satisfied. The reward function of the EI environment is defined as
\begin{equation}
	\label{eq-r_p}
	r_p^{t,i} = -(\mathcal{R}_{reward}^{t,i} + \mathcal{R}_{penalty}^{t,i}) , 
\end{equation}
where $\mathcal{R}_{reward}^{t,i} = \omega_{p1} DE^t + \omega_{p2} UC^t$ is the weighted sum of edge computing delay and usage charge and $\omega_{p1} + \omega_{p2} = 1$. $\mathcal{R}_{penalty}^{t,i} = \sum_{j=5}^{7} \mu_j \vartheta_j$ is used to punish $\mathbf{A_p}$ when its action cannot satisfy constraints $C5-C7$, where $\mu_j$ is a penalty factor (a sufficiently large number in this paper) and $\vartheta_j \in \lbrace 0,1 \rbrace$ is a binary variable. We set $\vartheta_j=0$ if $Cj$ is satisfied, and $\vartheta_j=1$, otherwise. Note that, we cannot obtain $r_p^{t,i}$  immediately when $i < N^t $. This is because VNF placement is a process of $N^t$ stages and we cannot obtain $DE^t$ and $UC^t$ before all VNFs $f_i^t \in F^t$ are placed. Thus, we devise a delayed update reward (DUR) scheme to update rewards $r_p^{t,1},...,r_p^{t,N^t-1}$. To be specific, we set $\mathcal{R}_{reward}^{t,i}=0$ when $i < N^t $. After all VNFs are placed, we have $i = N^t $ and we calculate $\mathcal{R}_{reward}^{t,N^t}$ by $\mathcal{R}_{reward}^{t,N^t} = \omega_{p1} DE^t + \omega_{p2} UC^t$. Then, we update the reward values by
\begin{equation}
	\label{eq-DUR}
	r_p^{t,i} = r_p^{t,i} - \mathcal{R}_{reward}^{t,N^t} , i=1,...,N^t-1.
\end{equation}
As mentioned before, the VNF placement process within $t$ consists of $N^t$ stages. At each stage, a VNF is selected to be placed on one BS. It is possible that the $i$-th VNF cannot be placed successfully, $i \leq N^t$. Simply setting all rewards $r_p^{t,i}$, $i=1,...,N^t$, to the same value may cause agent $\mathbf{A_p}$ to ignore some reasonable actions in the future. The DUR scheme avoids such situation. 

\subsection{TD3-based Task Partition}
As aforementioned, task partition is a continuous optimization problem, i.e., the action space of $\mathbf{A_o}$ is continuous. DDPG, a classic DRL algorithm, is widely used to solve problems with continuous action space \cite{lillicrap_ddpg_2015}. However, DDPG often suffers from overestimation of the Q-value and results in sub-optimal policy \cite{dong_deep_2020}. TD3 \cite{fujimoto2018addressing}, the direct successor of DDPG, overcomes the above problem by leveraging three novel mechanisms, namely clipped double Q-learning for actor-critic, target policy smoothing, and delayed policy updates.

This paper adapts TD3 to the task partition process. Different from DDPG that learns a single Q-function, TD3 adopts two critic networks and one actor network, as shown in Fig.~\ref{fig-framework}. The actor network consists of two subnets, i.e., one evaluate network $\pi_\phi$ with parameter $\phi$ and one target network $\pi_{\phi'}$ with parameter $\phi'$. The critic networks include two evaluate networks $Q_{\theta_1}$, $Q_{\theta_2}$ with parameters $\theta_1$, $\theta_2$, and two target networks $Q_{\theta_1'}$, $Q_{\theta_2'}$ with parameters $\theta_1'$, $\theta_2'$. 

\subsubsection{Clipped Double Q-Learning for Actor-Critic}
Evaluate networks $Q_{\theta_1}$ and $Q_{\theta_2}$ use a single target $y_o^j$, i.e., the smaller value output by target networks $Q_{\theta_1'}$ and $Q_{\theta_2'}$, defined as
\begin{equation}
	\label{eq-clip_dql}
	y_o^j = r_o^j + \gamma \underset{i=1,2}{\text{min}} Q_{\theta_i'} \left(  s_o^{j+1},\tilde{a}_o \right) ,
\end{equation}
where $\tilde{a}_o$ is the action generated by the target policy smoothing scheme, as defined in Eq.~\eqref{eq-smooth}. We update the parameters $\theta_i$ of $Q_{\theta_i}$, $i$ = 1, 2, by minimizing the loss function, defined as
\begin{equation}
	\label{eq-loss_critic}
	L(\theta_i) =  \frac{1}{N_b} \sum_{j=1}^{N_b} \left(  y_o^j - Q_{\theta_i'} (s_o^j, a_o^j ) \right) ^2 , i=1,2 ,
\end{equation}
where $N_b$ is the number of transitions in a mini-batch.

\subsubsection{Target Policy Smoothing}
The action, $\tilde{a}_o$, used to calculate target $y_o^j$ is obtained from the target action network $\pi_{\phi'}$, with clipped noise considered. After adding the noise, we also clip $\tilde{a}_o$ to lie in the valid action range $(a_o^{low},a_o^{high})$, ensuring $\tilde{a}_o \in \mathcal{A}_o $. Therefore, $\tilde{a}_o$ is defined as
\begin{equation}
	\label{eq-smooth}
	\tilde{a}_o = \text{clip}\left( \pi_{\phi'}(s_o^{j+1}) + \text{clip}(\epsilon,-c,c),a_o^{low},a_o^{high} \right) ,
\end{equation}
where $\epsilon \backsim \mathcal{N}(0,\tilde{\sigma})$ is the random noise defined by parameter $\tilde{\sigma}$, and $c>0$. According to the action space in Eq.~\eqref{eq-action_o}, we have $a_o^{low}=0$ and $a_o^{high}=1$. Here, the noise $\epsilon$ is a regularizer for TD3, helping it to achieve smoother state-action value estimation.

\begin{algorithm}[!t]
	\caption{TD3-based Task Partition}\label{alg:TD3}
	\begin{algorithmic}[1]
		\STATE Initialize evaluate critic networks $Q_{\theta_1}$ and $Q_{\theta_2}$, and evaluate actor network $\pi_\phi$ with random parameters $\theta_1$, $\theta_2$, and $\phi$;
		\STATE Initialize target networks $Q_{\theta_1'}$, $Q_{\theta_2'}$, $\pi_{\phi'}$ with $\theta_1' \leftarrow \theta_1$, $\theta_2' \leftarrow \theta_2$, $\phi' \leftarrow \phi$;
		\STATE Set replay buffer $\mathcal{B}_o$ to $ \emptyset$;
		\FOR{$k=1,...,EPS_{max}$}
		\STATE Observe state $s_o^1$;
		\FOR{$t=1,...,T$}
		\STATE Select action $a_o^t = \text{clip}(\pi_\phi(s_o^t) + \epsilon, 0, 1 )$;
		\STATE Take action $a_o^t$ and calculate local computing results $DL^t$ and $EL^t$;
		
		\STATE Send offloading ratio $x^t = a_o^t$ to \textit{\textbf{the EI environment}};
		\STATE \textbf{$\mathbf{A_p}$} places all VNFs in $F^t$ on BSs for edge computing purpose;
		\STATE Receive edge computing results $DE^t$, $EE^t$ and $UC^t$ from \textit{\textbf{the EI environment}};
		
		\STATE Based on the local and edge computing results, calculate $r_o^t$ and observe $s_o^{t+1}$;
		\STATE Store transition $(s_o^t, a_o^t, r_o^t, s_o^{t+1})$ in $\mathcal{B}_o$;
		\STATE Sample a mini-batch of $N_b$ transitions $(s_o^j, a_o^j, r_o^j, s_o^{j+1})$, $j = 1, ..., N_b$, from $\mathcal{B}_o$;
		\STATE Calculate $\tilde{a}_o$ according to Eq.~\eqref{eq-smooth};
		\STATE Set $y_o^j = r_o^j + \gamma \underset{i=1,2}{\text{min}} Q_{\theta_i'} (s_o^{j+1},\tilde{a}_o )$;			
		\STATE Update evaluate critic networks	\\
		$\theta_i \leftarrow \underset{\theta_i}{\text{min}} \frac{1}{N_b} \sum_{j} ( y_o^j - Q_{\theta_i'} (s_o^j, a_o^j ) )^2$, $i=1,2$ ;
		
		\IF{$(t \text{ mod } N_d) = 0$}
		{
			\STATE Update $\phi$ by the deterministic policy gradient:	\\
			$\nabla_\phi J(\phi) = \frac{1}{N_b} \sum_{j} [\nabla_{a_o^j} 
			\left. Q_{\theta_1} (s_o^j, a_o^j ) \right|_{a_o^j=\pi_\phi(s_o^j)} \cdot \nabla_\phi \pi_\phi(s_o^j)] $;
			\STATE Update target networks with \\
			$\theta_i' \leftarrow \tau \theta_i + (1-\tau) \theta_i'$,	$i=1,2$ ,	\\
			$\phi' \leftarrow \tau \phi + (1-\tau) \phi'$;
		}
		\ENDIF
		\ENDFOR
		\ENDFOR
	\end{algorithmic}
	\label{TD3}
\end{algorithm}

\subsubsection{Delayed Policy Updates}
TD3 updates evaluate actor network $\pi_\phi$ and target networks $Q_{\theta_1'}$, $Q_{\theta_2'}$ and $\pi_{\phi'}$ once, after the evaluate critic networks $Q_{\theta_1}$ and $Q_{\theta_2}$ are updated $N_d$ times, which is referred to as delayed policy updates \cite{fujimoto2018addressing}. That is, policy and target networks are not updated unless $Q_{\theta_1}$ and $Q_{\theta_2}$ are updated sufficiently. We use parameter $\theta_1$ of $Q_{\theta_1}$ to update $\pi_\phi$ via the policy gradient method as
\begin{equation}
	\label{eq-PG}
	\nabla_\phi J(\phi) = \frac{1}{N_b} \sum_{j} \nabla_{a_o^j} 
	\left. Q_{\theta_1} (s_o^j, a_o^j ) \right|_{a_o^j=\pi_\phi(s_o^j)} \nabla_\phi \pi_\phi(s_o^j).
\end{equation}
After that, we adopt soft update to update $\theta_1'$, $\theta_2'$, and $\phi'$ slowly, defined as
\begin{equation}
	\label{eq-update_targets}
	\begin{split}
		&\theta_i' \leftarrow \tau \theta_i + (1-\tau) \theta_i' ,	i=1,2,	\\
		&\phi' \leftarrow \tau \phi + (1-\tau) \phi',
	\end{split}
\end{equation}
where $\tau<<1$ is the learning rate.

The pseudo-code of TD3-based task partition is shown in Algorithm~\ref{alg:TD3}. At the beginning, agent $\mathbf{A_o}$ initializes the evaluate and target networks. The replay buffer of $\mathbf{A_o}$, $\mathcal{B}_o$, is set to $\emptyset$. Then, $\mathbf{A_o}$ starts interacting with the MD environment. In time slot $t$, $\mathbf{A_o}$ generates an action according to the current sate, defined as
\begin{equation}
	\label{eq-action_ao}
	a_o^t = \text{clip}(\pi_\phi(s_o^t) + \epsilon, 0, 1 ), 
\end{equation}
where $\epsilon \backsim \mathcal{N}(0,\sigma) $ is the exploration noise defined by parameter $\sigma$. As shown in Fig.~\ref{fig-framework}, right after action $a_o^t$ is generated, the MD environment sends offloading ratio $x^t = a_o^t$ to the EI environment and receives edge computing results $DE^t$, $EE^t$ and $UC^t$. Then, the MD environment transits to the next state $s_o^{t+1}$ and feeds back reward $r_o^t$ to $\mathbf{A_o}$. Transition $(s_o^t, a_o^t, r_o^t, s_o^{t+1})$ is added to the replay buffer $\mathcal{B}_o$.

In the training process, we randomly sample a mini-batch of $N_b$ transitions from $\mathcal{B}_o$. Based on target network $\pi_{\phi'}$ and state $s_o^{j+1}$, we generate action $\tilde{a}_o$ in favor of target policy smoothing. We use Eq.~\eqref{eq-clip_dql} to calculate target value $y_o^j$. After that, we update the two evaluate networks of critic $\theta_1$ and $\theta_2$ by minimizing the loss function defined in Eq.~\eqref{eq-loss_critic}. The deterministic policy gradient method in Eq.~\eqref{eq-PG} is adopted to update the evaluate actor network. Then, we soft update target networks $\theta_1'$, $\theta_2'$, and $\phi'$ according to Eq.~\eqref{eq-update_targets}. Finally, after training $\mathbf{A_o}$ $EPS_{max}$ rounds, we get the optimal policy for task partition.

\subsection{Dueling DDQN-based VNF Placement}
VNF placement is a discrete optimization problem, i.e., the action space of $\mathbf{A_p}$ is non-continuous. Model-free DRL approaches like DQN and DDQN are value-based algorithms \cite{van2016deep}, which have been widely applied to tackle such problems. For most DRL tasks, the values of a Q-function are distinct in different state-action pairs. Nevertheless, in some states, actions have no significant impact on the Q-function values. To solve this problem, dueling network architecture has been proposed in \cite{wang2016dueling}. The dueling network separates the representation of state values and action advantages, which can learn state values without influence of actions. Dueling DDQN proposed in \cite{hessel2018rainbow} combines the advantages of dueling network architecture and DDQN, making it suitable for the POSMJO problem formulated in this paper. Therefore, dueling DDQN is adapted to the VNF placement process.

As shown in Fig.~\ref{fig-framework}, the dueling DDQN consists of an evaluate network $Q_{\theta_p}$ with parameter $\theta_p$ and a target network $Q_{\theta_p'}$ with parameter $\theta_p'$. Dueling network architecture is adopted in both $Q_{\theta_p}$ and $Q_{\theta_p'}$. For example, the evaluate network, $Q_{\theta_p}$, consists of two streams that represent the value function $V_{\theta_p}$ and advantage function $A_{\theta_p}$, defined as
\begin{equation}
	\label{eq-dueling_1}
	Q_{\theta_p}(s_p^j,a_p^j) = V_{\theta_p}(s_p^j;\beta) + A_{\theta_p}(s_p^j,a_p^j;\alpha) ,
\end{equation}
where $\beta$ and $\alpha$ are the parameters of $V_{\theta_p}$ and $A_{\theta_p}$ respectively. $V_{\theta_p}$ is only related to states, while $A_{\theta_p}$ is related to both states and actions. Eq.~\eqref{eq-dueling_1} can be directly applied to get the state-action value. However, this equation is unidentifiable, e.g., given $Q_{\theta_p}$, we cannot recover $V_{\theta_p}$ and $A_{\theta_p}$ uniquely \cite{wang2016dueling}. In order to obtain identifiability, $Q_{\theta_p}$ is defined as 
\begin{equation}
	\label{eq-dueling_2}
	\begin{split}
		Q_{\theta_p} & (s_p^j,a_p^j) = V_{\theta_p}(s_p^j;\beta) + \\
		&\left(  A_{\theta_p}(s_p^j,a_p^j;\alpha)
		- \frac{1}{|\mathcal{A}_p|} \sum_{a_p\in \mathcal{A}_p} A_{\theta_p}(s_p^j,a_p;\alpha) \right)   ,
	\end{split}
\end{equation}
where $|\mathcal{A}_p|$ is the cardinality of $\mathcal{A}_p$.

In dueling DDQN, the loss function reflecting the estimation performance, is defined as 
\begin{equation}
	\label{eq-loss_ddqn}
	L(\theta_p) =  \mathbb{E}_{s_p^j,a_p^j,r_p^j,s_p^{j+1}} \left[  \left(  y_p^j - Q_{\theta_p} (s_p^j, a_p^j ) \right) ^2 \right] , 
\end{equation}
where $y_p^j$ is the target defined as 
\begin{equation}
	\label{eq-target_ddqn}
	y_p^j = r_p^j + \gamma Q_{\theta_p'} \left(  s_p^{j+1}, \underset{a_p}{\text{argmax}} Q_{\theta_p}(s_p^{j+1}, a_p) \right)  . 
\end{equation}
The loss function in Eq.~\eqref{eq-loss_ddqn} can be minimized by stochastic gradient descent for updating parameter $\theta_p$ \cite{van2016deep}.

The pseudo-code of dueling DDQN-based VNF placement is shown in Algorithm~\ref{alg:D3QN}. 
At the beginning, agent $\mathbf{A_p}$ initializes the evaluate and target networks, respectively. Different from $\mathbf{A_o}$, $\mathbf{A_p}$ is associated with two replay buffers, i.e., $\mathcal{B}_p$ and $\mathcal{B}_p^t$. $\mathcal{B}_p$ is used to store transitions during training, while $\mathcal{B}_p^t$ is adopted to store transitions corresponding to the VNF placement decisions of $u^t$ in time slot $t$. After $Q_{\theta_p}$, $Q_{\theta_p'}$ and $\mathcal{B}_p$ are initialized, $\mathbf{A_p}$ starts interacting with the EI environment. 

Recall that, within $t$, the VNF placement of $u^t$ is divided into $N^t$ stages. 
Within $t$, $\mathbf{A_p}$ receives the offloading ratio, $x^t$, from the MD environment. Then, at stage $i$, $\mathbf{A_p}$ decides where to place the $i$-th VNF in $F^t$, $i=1,...,N^t$. Due to lack of experience in the early process, $\mathbf{A_p}$ generates an action by an $\varepsilon$-greedy strategy, defined as
\begin{equation}
	\label{eq-dqn_greedy}
	a_p^{t,i} = 
	\begin{cases} 
		\text{a random action}, & \text {with probability $\varepsilon$} 
		\\ \underset{a_p}{\text{argmax}} Q_{\theta_p}(s_p^{t,i},a_p ), & \text{wth probability $1-\varepsilon$}
	\end{cases}
\end{equation}
Subsequently, the EI environment transits to the next state $s_p^{t,i+1}$ and feeds back reward $r_p^{t,i}$ to $\mathbf{A_p}$. Transitions $(s_p^{t,i}, a_p^{t,i}, r_p^{t,i}, s_p^{t,i+1})$ is added to the replay buffer $\mathcal{B}_p^t$. After all VNFs in $F^t$ are placed, the DUR scheme is adopted to update the rewards of transitions in $\mathcal{B}_p^t$ according to Eq.~\eqref{eq-DUR}. After DUR, all transitions in $\mathcal{B}_p^t$ are stored in $\mathcal{B}_p$. Then, edge computing results, $DE^t$, $EE^t$ and $UC^t$, are sent to the MD environment.

In the training process, we randomly sample a mini-batch of $N_b$ transitions from $\mathcal{B}_p$. We calculate the target value $y_p^j$ according to Eq.~\eqref{eq-target_ddqn}. Gradient descent is performed, with the loss function defined in Eq.~\eqref{eq-loss_ddqn}. After that, we update the target network parameter $\theta_p'$ every $N_d$ steps. Finally, after training $\mathbf{A_p}$ $EPS_{max}$ rounds, we get the optimal policy for VNF placement.

\begin{algorithm}[!t]
	\caption{Dueling DDQN-based VNF Placement}\label{alg:D3QN}
	\begin{algorithmic}[1]
		\STATE Initialize evaluate network $Q_{\theta_p}$ with random parameters $\theta_p$;
		\STATE Initialize target network $Q_{\theta_p'}$ with $\theta_p' \leftarrow \theta_p$;
		\STATE Set replay buffer $\mathcal{B}_p$ to $ \emptyset$;
		\FOR{$k=1,...,EPS_{max}$}
		\FOR{$t=1,...,T$}
		\STATE  Receive offloading ratio $x^t$ from \emph{\textbf{the MD environment} };
		\STATE Observe state $s_p^{t,1}$;
		\STATE Set $\mathcal{B}_p^t=\emptyset$;
		\FOR{$i=1,...,N^t$}
		\STATE Select $a_p^{t,i}$ according to Eq.~\eqref{eq-dqn_greedy};	// $\varepsilon$-greedy
		\STATE Take action $a_p^{t,i}$, calculate $r_p^{t,i}$ and observe $s_p^{t,i+1}$;
		\STATE Store transition $(s_p^{t,i}, a_p^{t,i}, r_p^{t,i}, s_p^{t,i+1})$ in $\mathcal{B}_p^t$;
		\IF{every $f_i^t \in F^t$ is placed}	
		{
			\STATE Calculate $\mathcal{R}_{penalty}^{t,i}$, $i=1,..,N^t$;
			\STATE Update transitions in $\mathcal{B}_p^t$ according to Eq.~\eqref{eq-DUR};
			\STATE Set $\mathcal{B}_p = \mathcal{B}_p \cup \mathcal{B}_p^t $;
		}
		\ENDIF
		\ENDFOR
		
		\STATE Send edge computing results $DE^t$, $EE^t$ and $UC^t$ to \emph{\textbf{the MD environment} };
		
		\STATE Sample a mini-batch of $N_b$ transitions $(s_p^j, a_p^j, r_p^j, s_p^{j+1})$, $j = 1, ..., N_b$, from $\mathcal{B}_p$;
		\IF{transition is terminal}
		\STATE Set $y_p^j = r_p^j$;
		\ELSE
		\STATE Set $y_p^j = r_p^j + \gamma Q_{\theta_p'} ( s_p^{j+1}, \underset{a_p}{\text{argmax}} Q_{\theta_p}(s_p^{j+1}, a_p) )$;
		\ENDIF
		\STATE Perform a gradient descent step on	\\
		$(y_p^j - Q_{\theta_p}(s_p^{j}, a_p^j) )^2$;
		\STATE Update target network every $N_d$ steps, i.e., $\theta_p' \leftarrow \theta_p$;
		
		\ENDFOR
		\ENDFOR
		
	\end{algorithmic}
	\label{D3QN}
\end{algorithm}

\subsection{Cooperative Dual-agent DRL Algorithm}
\begin{figure}[!t]
	\centering
	\includegraphics[width=3.5in]{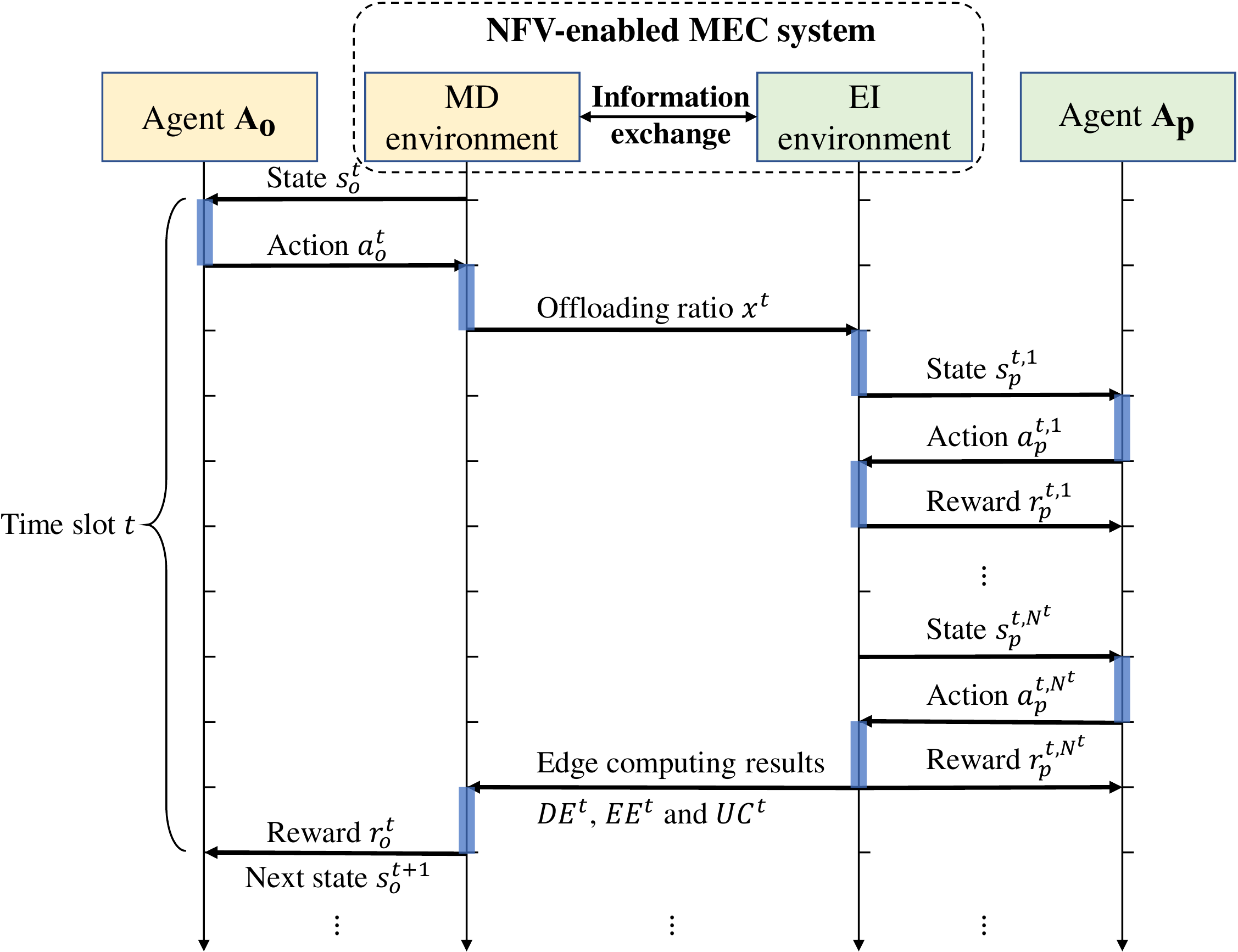}
	\caption{Swimlane diagram of the interaction between agents and environments.}
	\label{fig-swimlane}
\end{figure}

In the CDADRL algorithm, agents $\mathbf{A_o}$ and $\mathbf{A_p}$ collaborate on tackling the POSMJO problem. The state, $s_p^t$ of $\mathbf{A_p}$, depends on the offloading ratio $x^t = a_o^t$ made by $\mathbf{A_o}$, and the reward, $r_o^t$ of $\mathbf{A_o}$, is calculated based on the edge computing results corresponding to the VNF placement solution made by $\mathbf{A_p}$. 
As shown in Fig.~\ref{fig-framework}, the collaboration between the two agents is realized by the information exchange between the MD and EI environments. The swimlane diagram shown in Fig.~\ref{fig-swimlane} depicts the interaction process between the agents and environments. To be specific, between $\mathbf{A_o}$ and the MD environment, $\mathbf{A_o}$ observes the current state, $s_o^t$, and takes action ,$a_o^t$. After receving the edge computing results from the EI environment, $DE^t$, $EE^t$ and $UC^t$, the MD environment feeds back reward $r_o^t$ to $\mathbf{A_o}$ and transits to the next state, $s_o^{t+1}$. Similarly, after the MD environment receives the offloading ratio $x^t = a_o^t$ from the EI environment, $\mathbf{A_p}$ observes the current state, $s_p^{t,i}$, and takes action, $a_p^{t,i}$. Then, the EI environment feeds back reward $r_p^{t,i}$ to $\mathbf{A_p}$ and transits to the next state $s_p^{t,i+1}$.

\begin{figure*}[!t]
	\centering
	\includegraphics[width=7in]{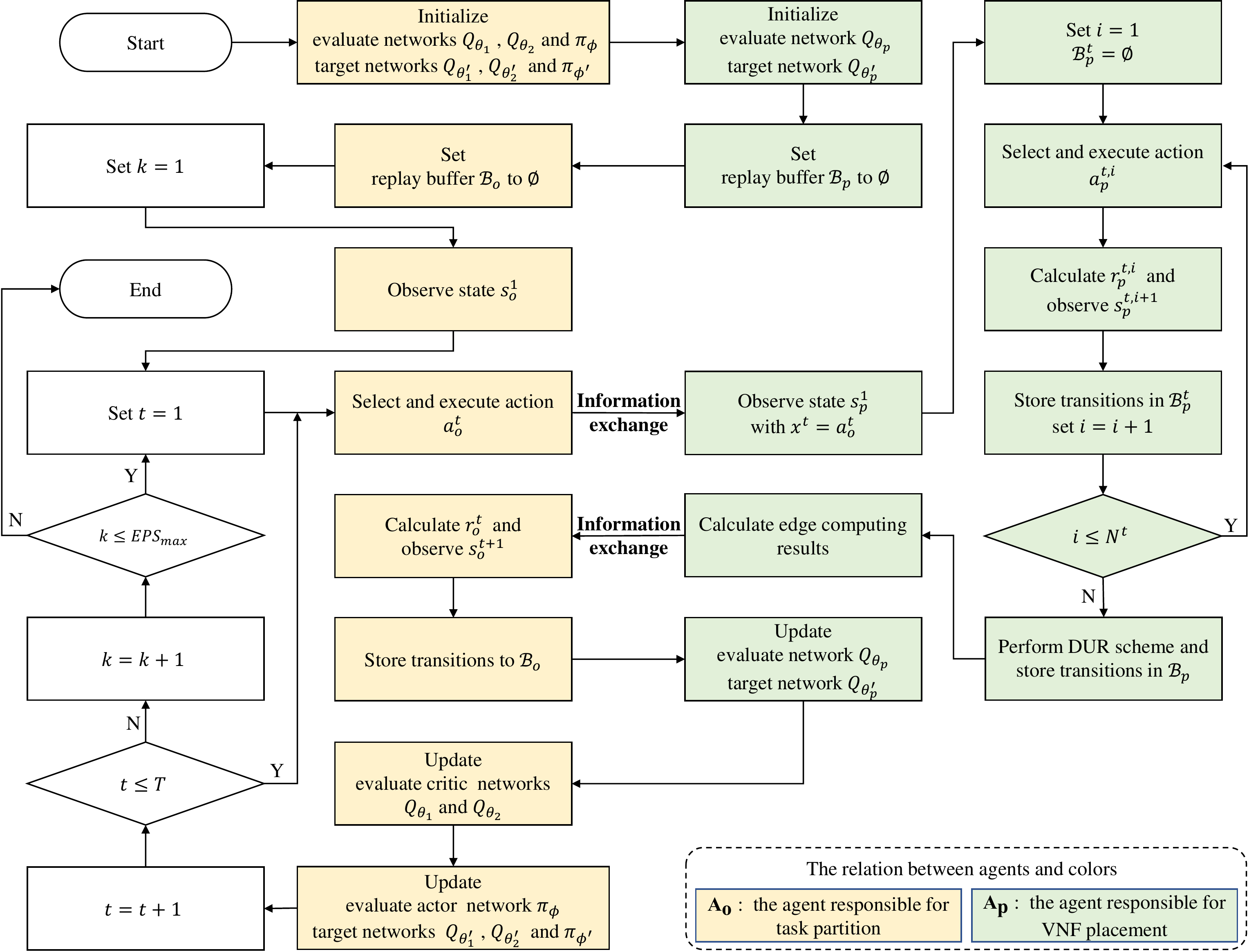}
	\caption{Flowchart of the proposed CDADRL algorithm.}
	\label{fig-flowchart}
\end{figure*}

The flowchart of the proposed CDADRL algorithm is shown in Fig.~\ref{fig-flowchart}. It can be seen that, $\mathbf{A_o}$ and $\mathbf{A_p}$ interact with each other via the information exchange process between two environments. The two agents realize the two related decision-making steps through close cooperation. At the beginning, $\mathbf{A_o}$ and $\mathbf{A_p}$ initialize their neural networks and replay buffers respectively. In time slot $t$, $\mathbf{A_o}$ decides the offloading ratio $x^t$ first. After receiving $x^t$, $\mathbf{A_p}$ takes a series of actions to generate a VNF placement solution and stores transitions to $\mathcal{B}_p$. Then, $\mathbf{A_o}$ receives edge computing results and stores transition in $\mathcal{B}_o$. After that, $\mathbf{A_o}$ and $\mathbf{A_p}$ update their neural networks respectively. After training $\mathbf{A_o}$ and $\mathbf{A_p}$ $EPS_{max}$ rounds, we can get the optimal policies for task partition and VNF placement.

\section{ Simulation Results and Discussion} \label{Simulation Results and Discussion}

In this section, we evaluate the performance of the proposed CDADRL algorithm for the POSMJO problem. A python simulator running on Google Tensorflow 2.2 is developed for performance evaluation. In the simulation, we assume that the number of time slots is 20, i.e., $T=20$, and the duration of each time slot is 5 seconds. We consider the edge infrastructure consists of 10 BSs, where each BS is co-located with an MEC server and provides edge computing service to the MD. The distance from the MD to one BS is randomly generated in the range of [100, 800] m. 
The computing capacity of each BS, $cp_v$, is randomly generated in the range of [2, 6] GHz. The bandwidth capacity of each link, $bw_e$, is randomly generated in the range of [20, 100] Mbps. In each time slot, the input data size of the required task, $d$, is randomly generated in the range of [800, 1000] Kbits. 
In CDADRL, the neural network architecture used in all Q-networks is identical. This architecture is a four-layer full-connected network, with 64 neurons in each layer and rectified linear unit (ReLU) as the activation function.  The Adam optimizer is adopted for neural network training. The parameter of the $\varepsilon$-greedy exploration, $\varepsilon$, is annealed from 1 to 0.01, with a decay rate of 0.9995. Other parameter values used in the experiment are listed in Table~\ref{tab:parameters}. 
Unless specified, the parameter settings above are adopted for performance evaluation and comparison. The results are obtained by running each algorithm 10 times on a machine with Intel(R) Xeon(R) CPU E5-2667 v4 @ 3.20GHz and NVIDIA TITAN V GPU, from which the statistics are collected and analyzed.

\begin{table}[!t]
	\caption{Experimental Configuration.\label{tab:parameters}}
	\centering
	\begin{tabular}{ll}
		\hline
		Parameter & Value\\
		\hline
		Number of BSs & 10 \\
		Wireless bandwidth ($W$) & $20$ MHz\\
		Noise power ($\eta^2$) & $10^{-6}$ Watt\\
		Computing capacity of the MD ($cp_{MD}$) & $0.6$ GHz\\ 
		Transmission power of the MD  ($p_{MD}^{tr}$) & $0.5$ Watt\\
		Receiving power of the MD  ($p_{MD}^{re}$) & $0.1$ Watt\\
		Transmission power of each BS  ($p_v^{tr}$) & $[1,2]$ Watt\\
		
		Computing capacity of each BS ($cp_v$) & $[2,6]$ GHz\\
		Bandwidth capacity of each link ($bw_e$) & $[20,100]$ Mbps\\
		
		Input data size of each task ($d$) & $[800,1000]$ Kb\\
		Number of CPU cycles ($c$) & $[800,1000]$ cycle/bit\\
		Number of VNFs in an SFC & $[3,5]$ \\
		Required computing capacity of each VNF		 ($cp(f)$) & $[0.1,0.5]$ GHz\\
		Required bandwidth between any two VNFs ($br_i$) & $[5,10]$ Mbps\\
		Ratio of $h_i^t$ to $d^t$ ($\xi_i$) & $[0.5,1.5]$\\
		Execution deadline ($D_{max}$)  & $[20,35]$ Sec.\\
		Effective capacitance coefficient ($\kappa$) & $10^{-26}$\\
		
		Discount factor ($\gamma$) & 0.99\\
		Maximum number of episodes ($EPS_{max}$) & 3000\\
		\hline
	\end{tabular}
\end{table}

\subsection{Parameter Study of CDADRL Algorithm}
\begin{figure*}[!t]
	\centering
	\subfloat[Different learning rates]{\includegraphics[scale=0.27]{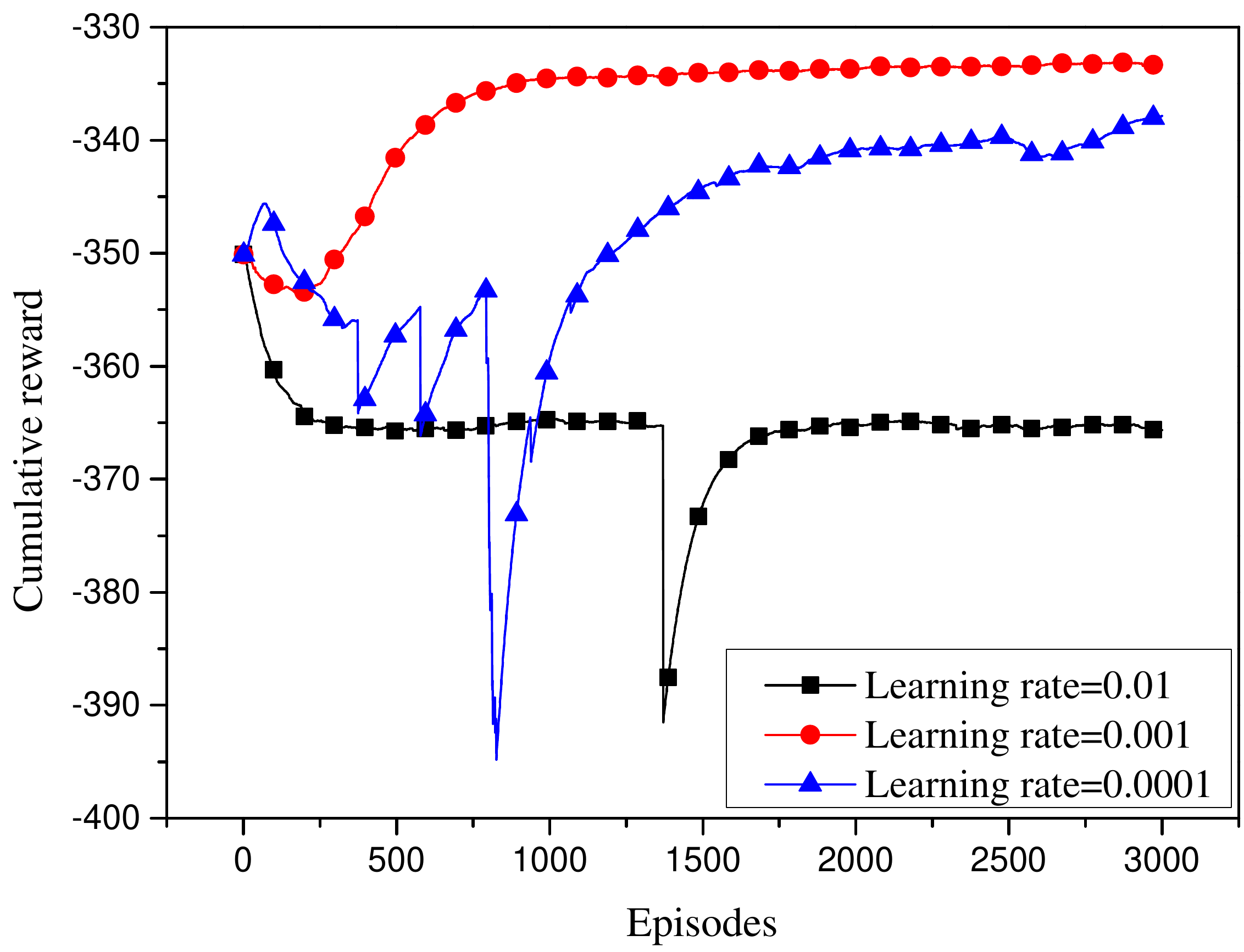}} 
	\subfloat[Different batch sizes]{\includegraphics[scale=0.27]{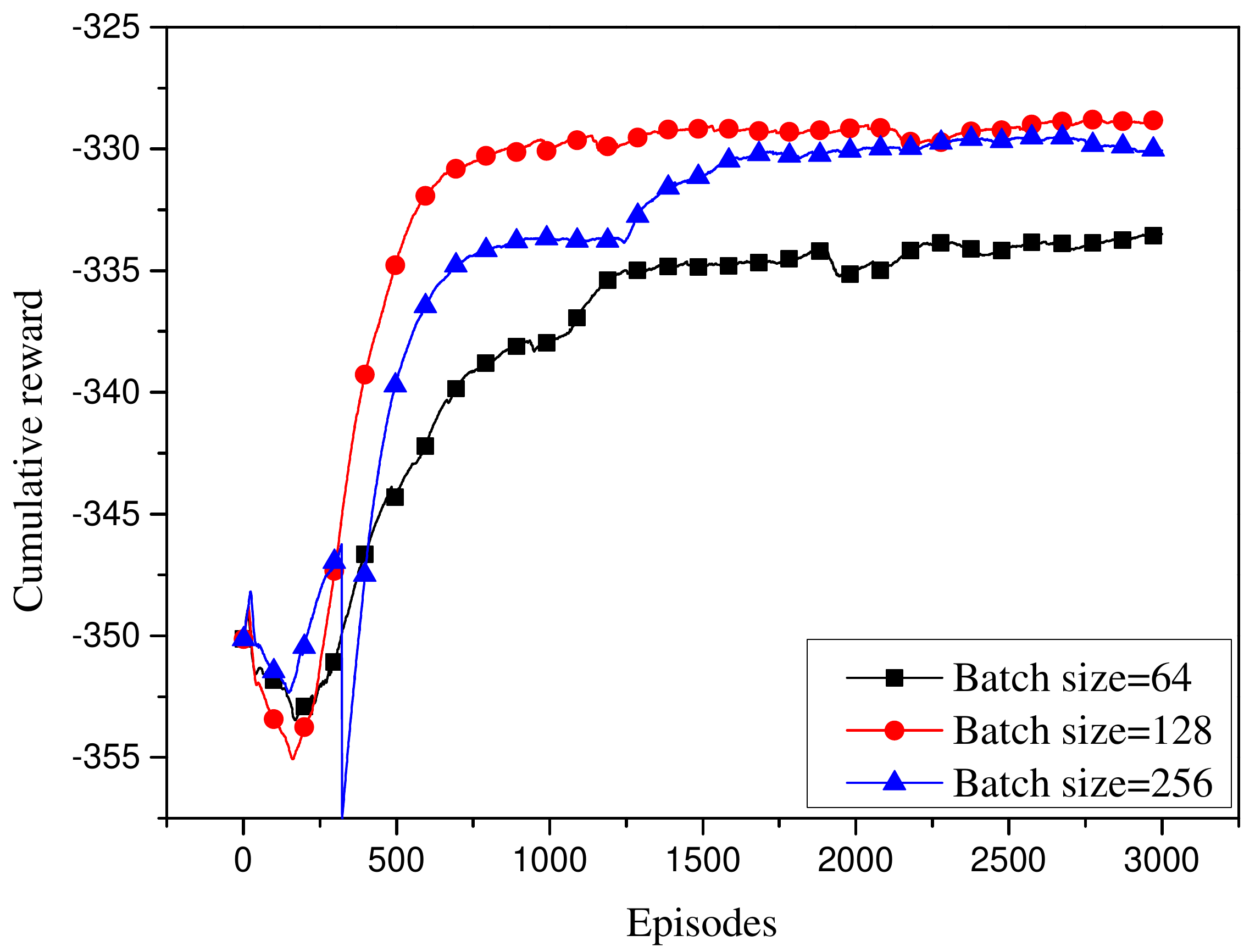}} 
	\subfloat[Different buffer sizes]{\includegraphics[scale=0.27]{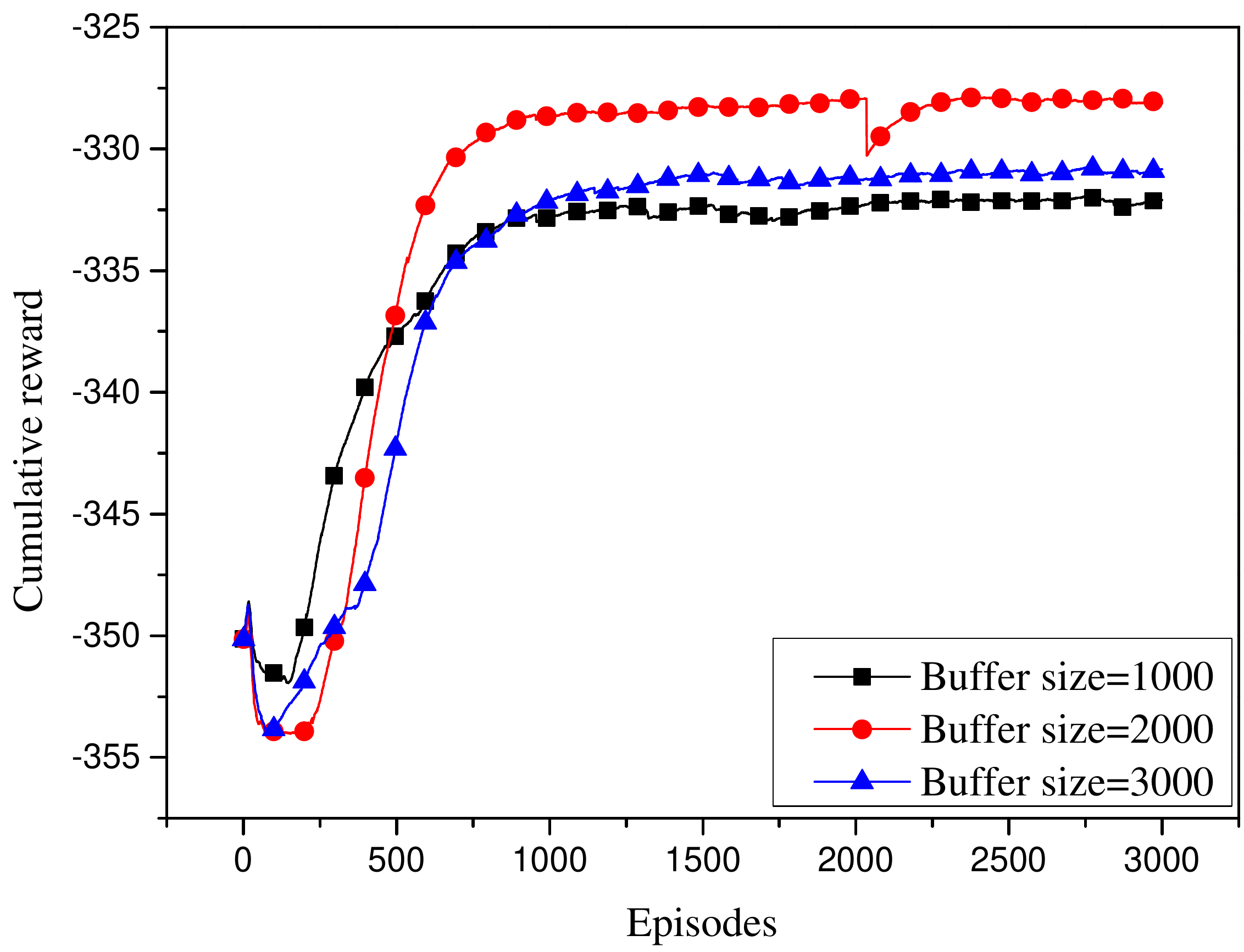}} 
	\caption{Cumulative reward of CDADRL algorithm with different parameters.}
	\label{fig-parameterStudy}
\end{figure*}

We study the impact of parameters on the performance of the proposed CDADRL algorithm, including the learning rate, batch size, and buffer size. The average cumulative reward of CDADRL algorithm is depicted in Fig.~\ref{fig-parameterStudy}. 
Fig.~\ref{fig-parameterStudy}(a) shows the cumulative rewards with different learning rates. As can be seen, both 0.0001 and 0.01 lead to poor training performance, and the cumulative reward curve oscillates at low values. This is because a too large learning rate leads to rapid changes to the parameters of neural networks, which makes the training process become unstable. A too small learning rate results in trivial changes to the neural networks' parameters, which leads to painfully slow convergence. Obviously, 0.001 is more appropriate than 0.01 and 0.0001. Thus, we hereafter fix the learning rate to 0.001. 
Fig.~\ref{fig-parameterStudy}(b) shows the cumulative rewards with different batch sizes. With a small batch size, i.e., 64, the algorithm cannot reach to high cumulative reward values because the gradients from a small batch may not  cover promising directions of optimization. On the contrary, a too large batch size, i.e., 256, may lead to frequent use of non-effective experiences. Therefore, we hereafter set the batch size to 128.
Fig.~\ref{fig-parameterStudy}(c) depicts the cumulative rewards with different buffer sizes. Clearly, the algorithm with a buffer size of 2000 obtains the best performance on training. This is because a smaller buffer stores less experiences for policy optimization and it is less likely to keep promising ones. A larger buffer maintains more promising experiences. However, out-of-date transitions are more likely to be picked up, disturbing the current direction of optimization. Hence, we hereafter set the buffer size to 2000.

\begin{figure*}[!t]
	\centering
	\subfloat[$cp_{MD}=0.6$]{\includegraphics[scale=0.27]{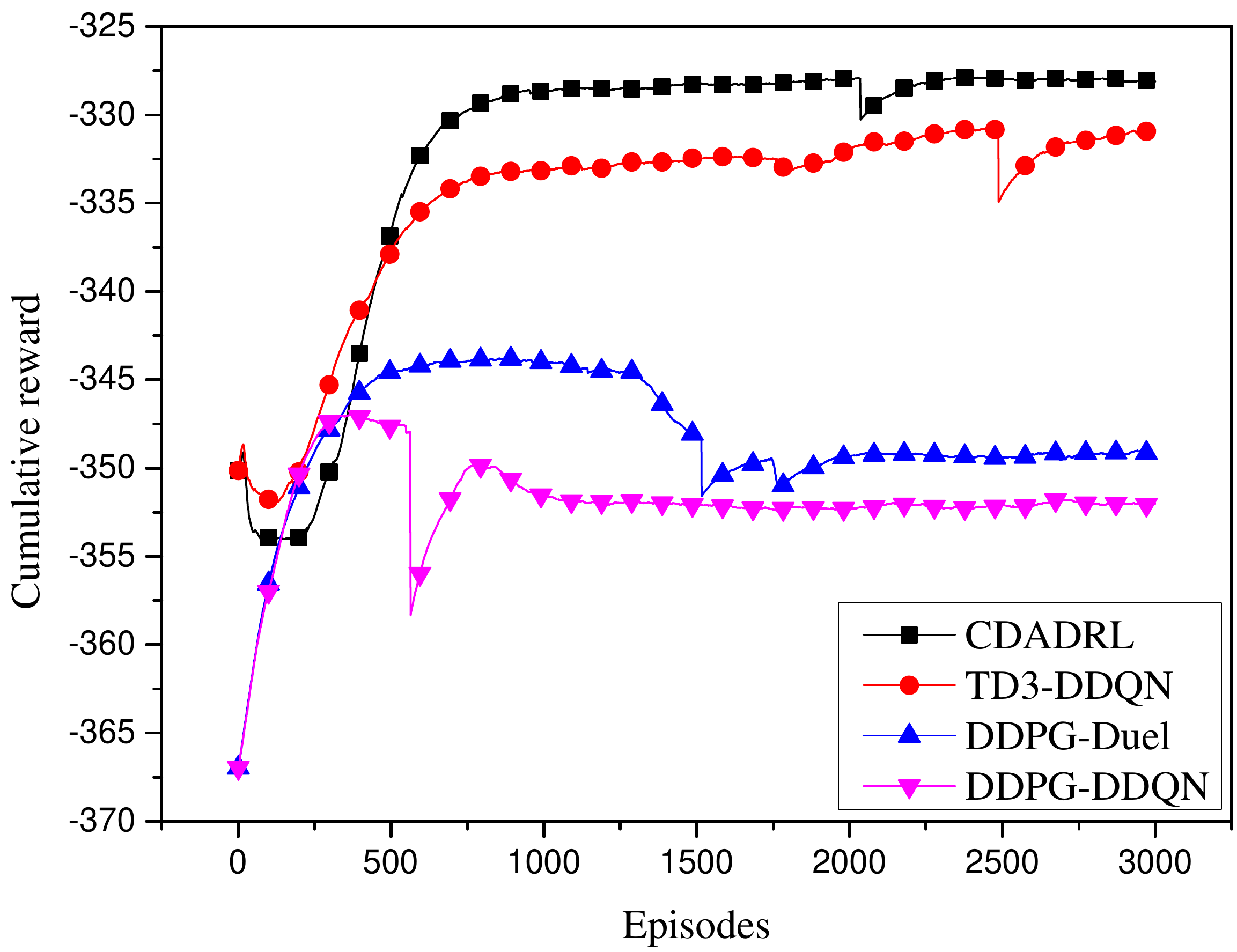}} 
	\subfloat[$cp_{MD}=0.8$]{\includegraphics[scale=0.27]{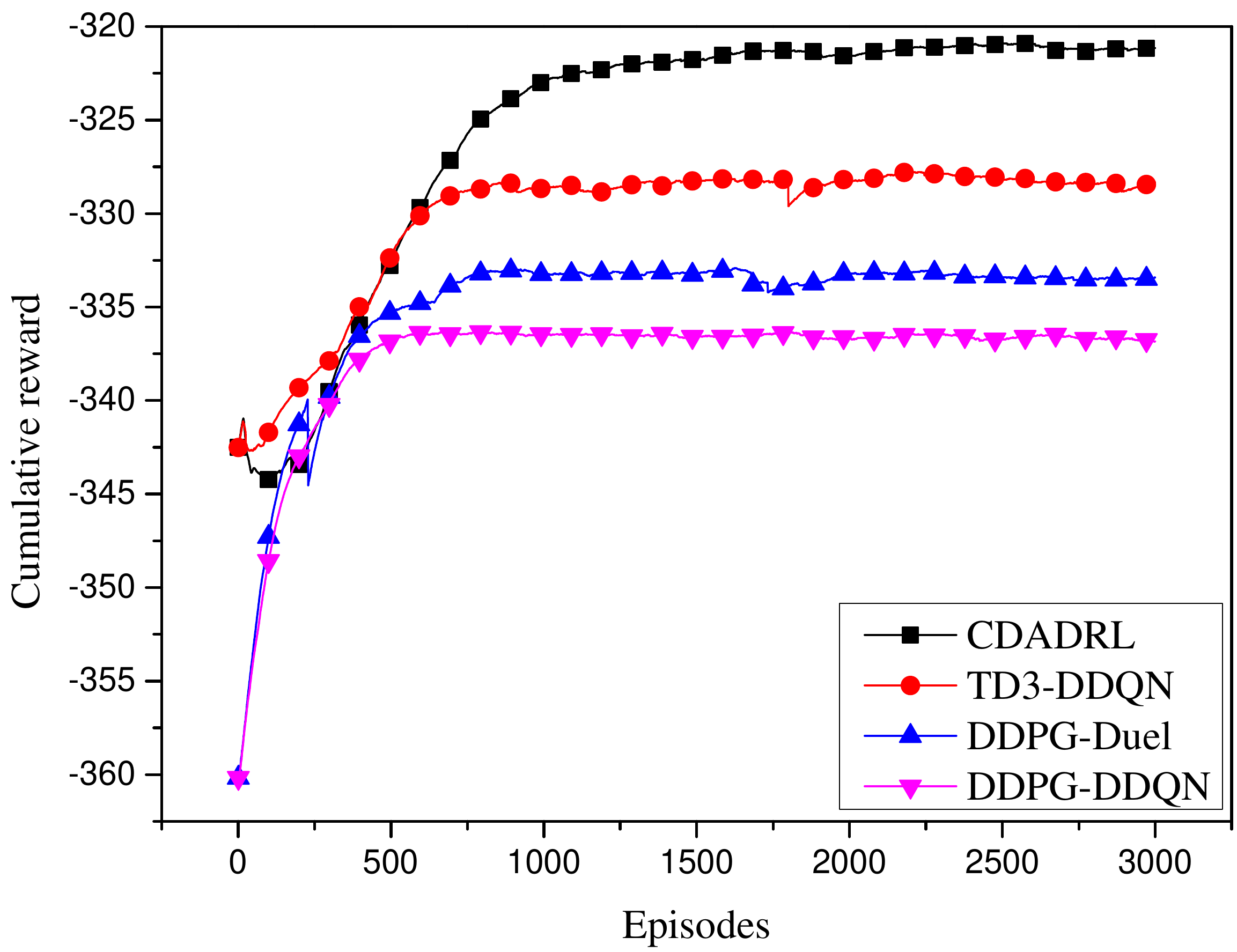}} 
	\subfloat[$cp_{MD}=1.0$]{\includegraphics[scale=0.27]{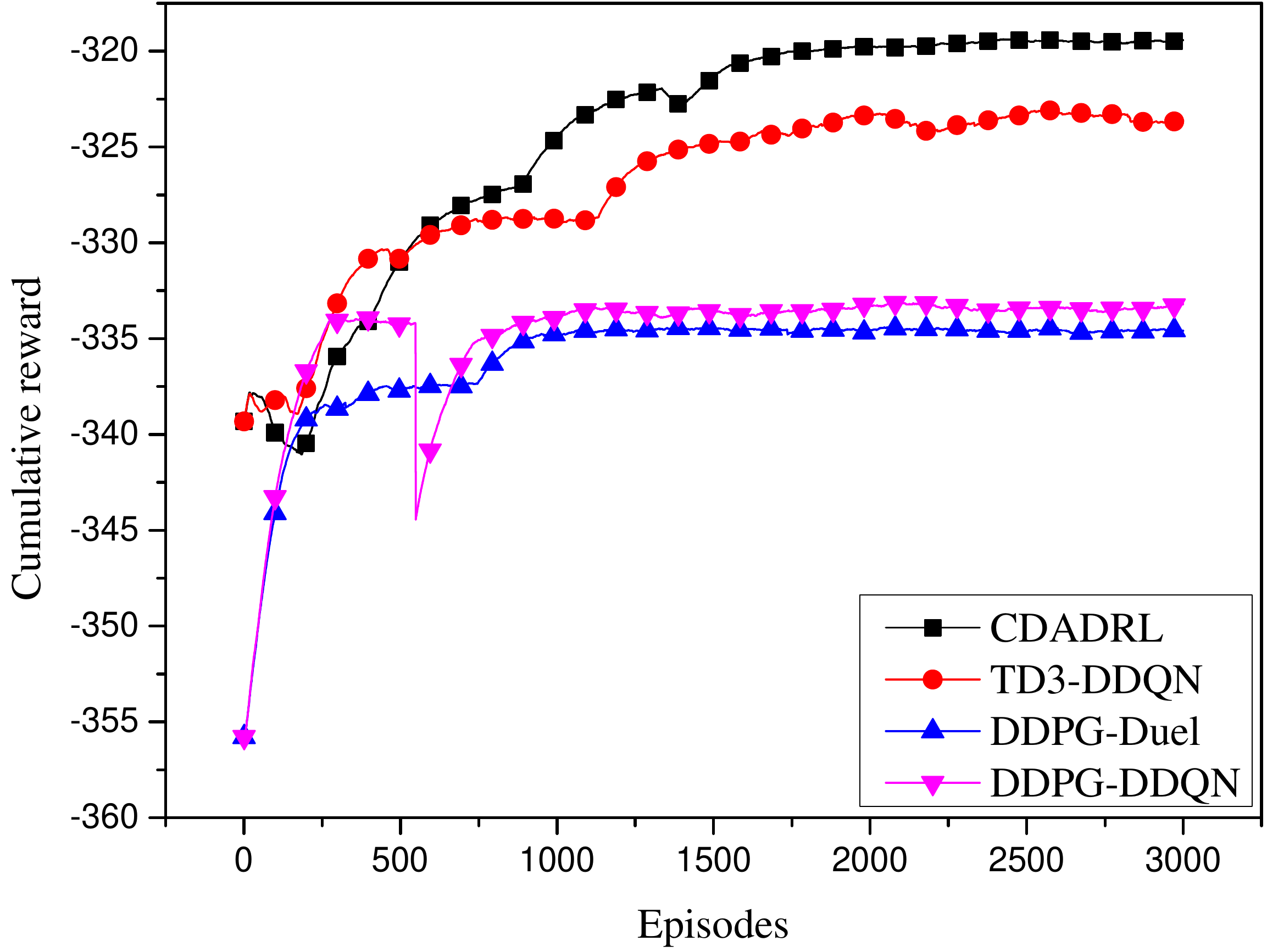}} \\
	\subfloat[$cp_{MD}=1.2$]{\includegraphics[scale=0.27]{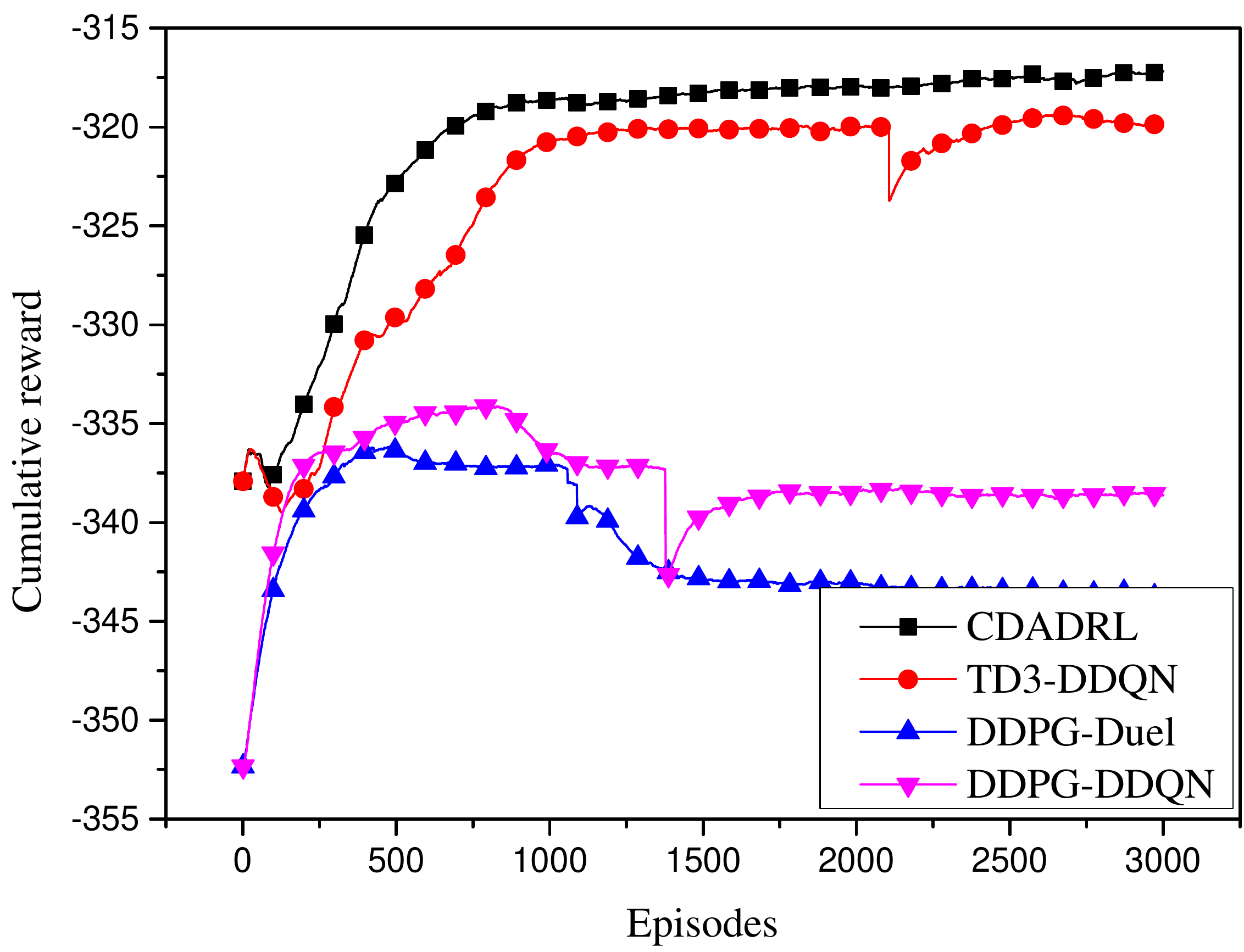}} 
	\subfloat[$cp_{MD}=1.4$]{\includegraphics[scale=0.27]{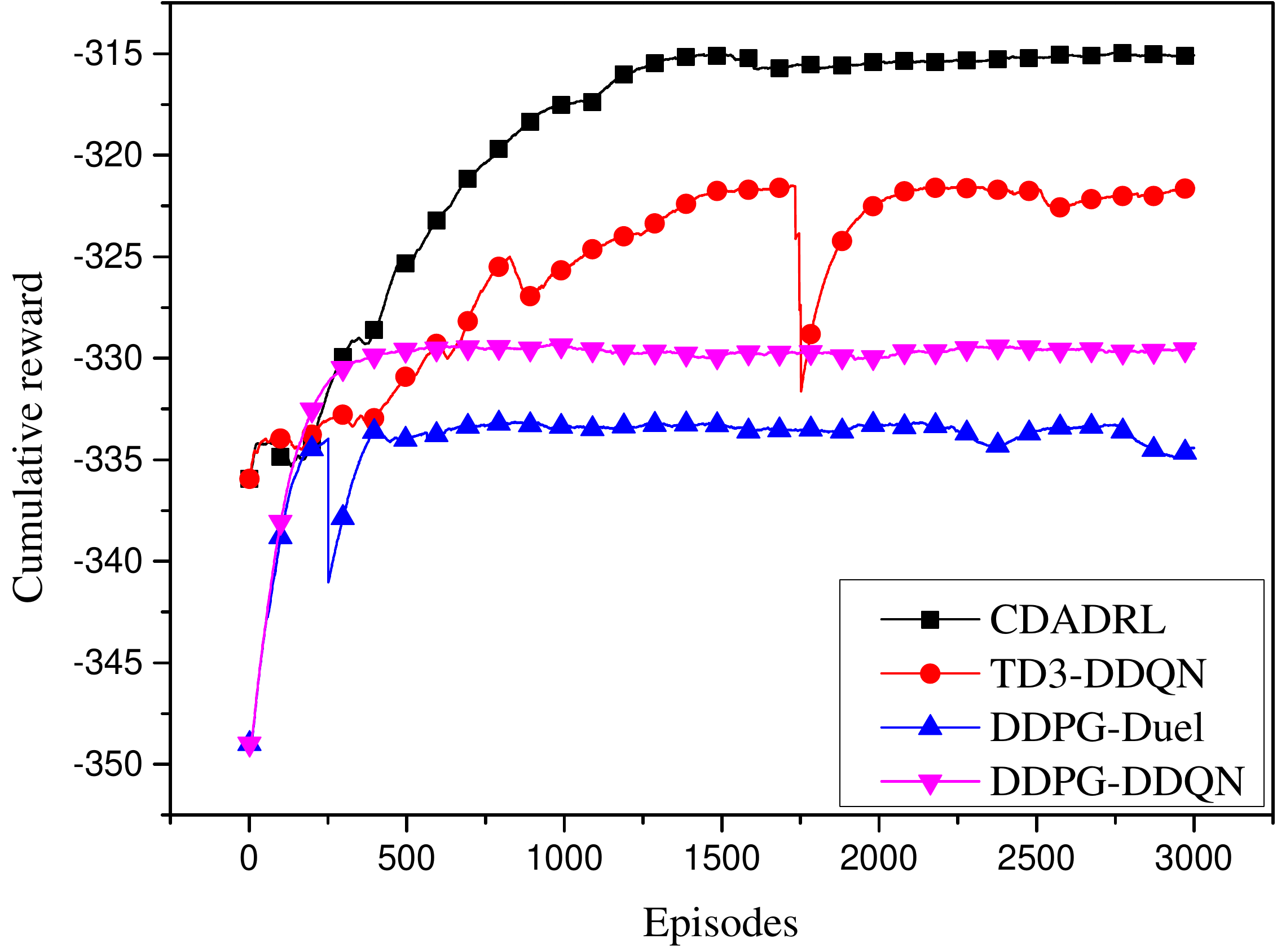}} 
	\subfloat[$cp_{MD}=1.6$]{\includegraphics[scale=0.27]{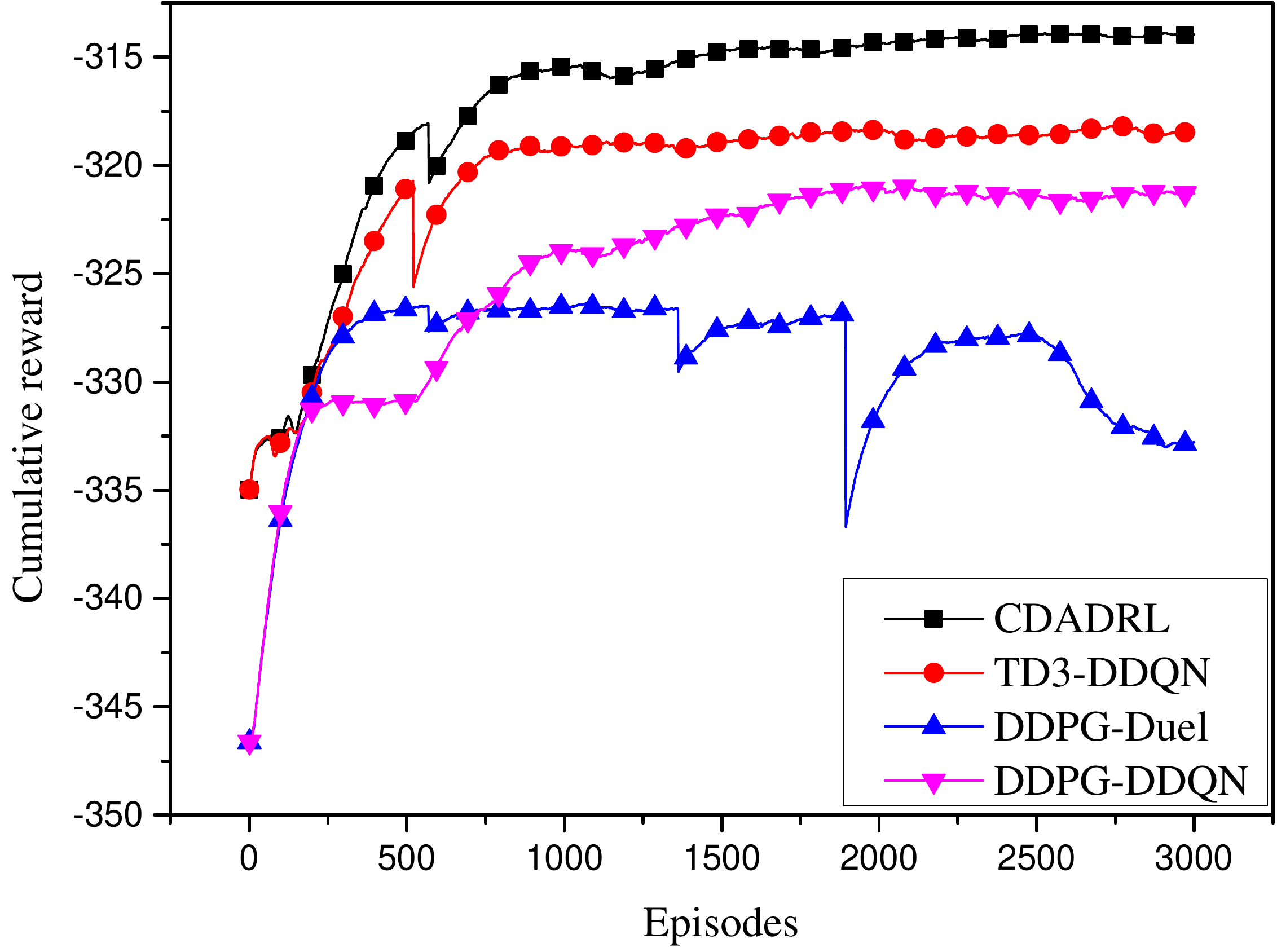}} 
	\caption{Cumulative reward of the four algorithms with different computing capacities of the MD (GHz).}
	\label{fig-DRLcomp}
\end{figure*}

\subsection{Overall Performance Evaluation}
Note that the POSMJO problem is brand-new and no existing algorithm is immediately available to tackle it. In order to evaluate the learning performance of the CDADRL algorithm, we compare it with three combinations of state-of-the-art DRL algorithms for continuous and discrete decision-making, against cumulative reward and average episodic reward. These combinations are TD3-DDQN, DDPG-Duel, and DDPG-DDQN, listed below.
\begin{itemize}
	\item \textbf{TD3-DDQN:} the original CDADRL algorithm with $\mathbf{A_p}$ based on the conventional DDQN \cite{van2016deep} instead of dueling DDQN.
	
	\item  \textbf{DDPG-Duel:} the original CDADRL algorithm with $\mathbf{A_o}$ based on DDPG \cite{lillicrap_ddpg_2015} instead of TD3.
	
	\item \textbf{DDPG-DDQN:} the original CDADRL algorithm with $\mathbf{A_o}$ based on DDPG and $\mathbf{A_p}$ based on DDQN.
	
	\item \textbf{CDADRL:} The cooperative dual-agent DRL algorithm proposed in this paper.
	
\end{itemize}

\begin{figure}[!t]
	\centering
	\includegraphics[width=3.2in]{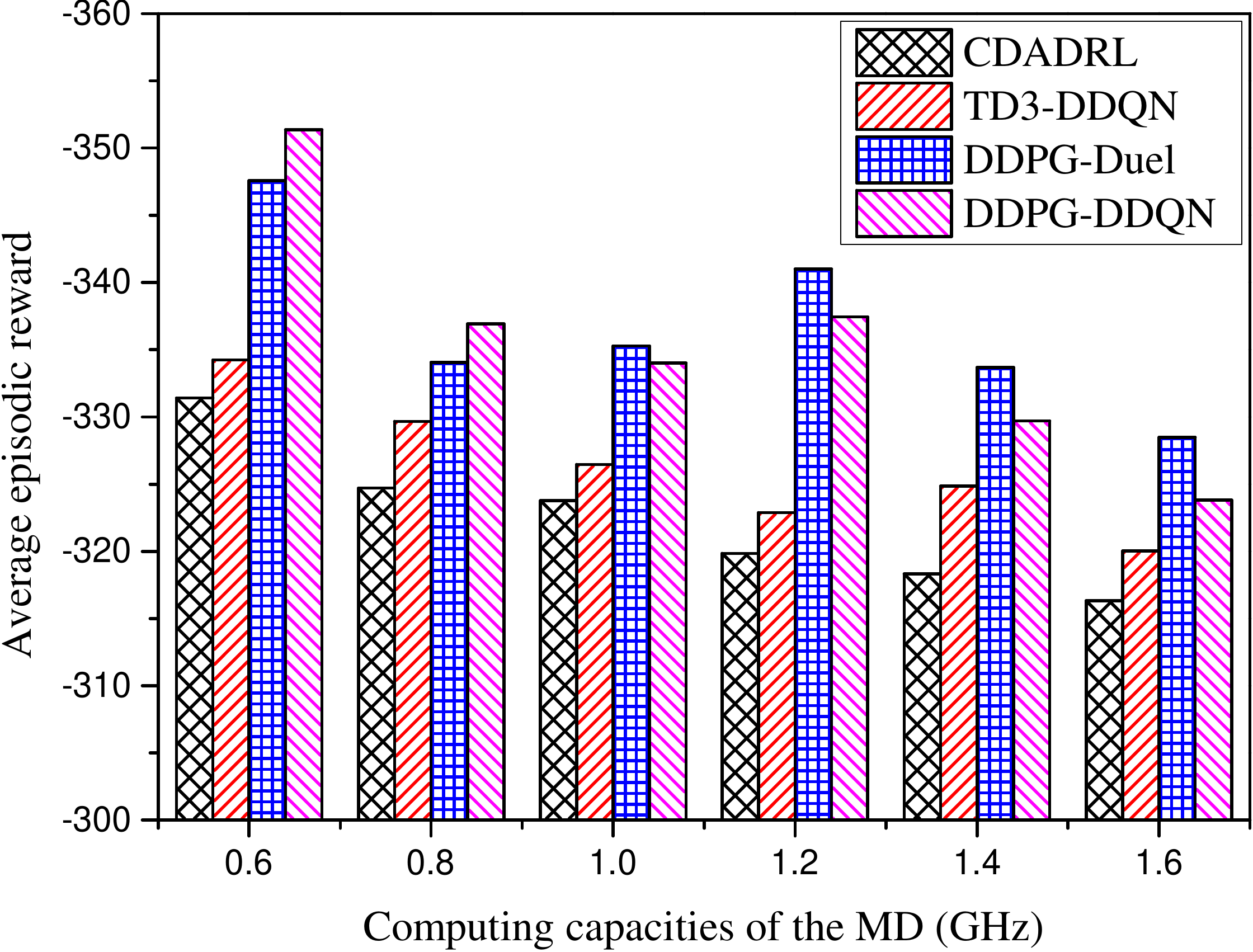}
	\caption{Results of average episodic reward.}
	\label{fig-AER}
\end{figure}

Fig.~\ref{fig-DRLcomp} shows the cumulative reward curves of the four algorithms with different computing capacities of the MD. It is easily seen that \textbf{CDADRL} and \textbf{TD3-DDQN} perform the best and second-best among the four algorithms. Compared to \textbf{TD3-DDQN}, \textbf{CDADRL} has a faster training speed in Figs.~\ref{fig-DRLcomp}(d), \ref{fig-DRLcomp}(e) and \ref{fig-DRLcomp}(f) and higher cumultive reward values once converged in the six cases. This is because dueling \textbf{DDQN} helps \textbf{CDADRL} to make better use of the state value, not just the state-action value, which is in favor of making more appropriate VNF placement decisions. Besides, \textbf{DDPG-Duel} and \textbf{DDPG-DDQN} are the two worst algorithms as they have poor performance on convergence. For example, the \textbf{DDPG-Duel} algorithm does not well converge after 3000 episodes when $cp_{MD}=1.6$. The reason behind this is the task partition step of \textbf{DDPG-Duel} and \textbf{DDPG-DDQN} is based on DDPG. DDPG is prone to overestimating Q-values, making it easily trapped into sub-optimal policies \cite{dong_deep_2020}. On the other hand, \textbf{CDADRL} adopts TD3 which leverages three novel mechanisms to overcome the above difficulty. Thus the combination of TD3 and Dueling DDQN is the most appropriate for addressing the POSMJO problem. The average episodic reward values obtained by the four algorithms are illustrated in Fig.~\ref{fig-AER}. Clearly, \textbf{CDADRL} obtains the smallest value in each case among the four algorithms. This also justifies why $\mathbf{A_o}$ and $\mathbf{A_p}$ are based on TD3 and dueling DDQN.

We also evaluate the offloading performance by comparing \textbf{CDADRL} with four baseline offloading schemes, in terms of execution delay, energy consumption of the MD, and usage charge for edge computing. The four algorithms with different offloading schemes are listed below.

\begin{itemize}
	\item  \textbf{Local:} Any incoming task is executed on the MD, which means no task is to be offloaded to the edge.
	
	\item \textbf{Edge:} All incoming tasks are offloaded to the MEC servers for processing, where the DQN-based online SFC deployment method is adopted to place VNFs on MEC servers \cite{xu2022cloud}.
	
	\item \textbf{Binary:} An incoming task is either executed locally on the MD or offloaded completely to the edge for processing. Assume a task is executed locally with a probability of $50\%$. The online algorithm proposed in \cite{xu2019task} is applied to VNF placement.
	
	\item \textbf{Random:} For an arbitrary incoming task, its offloading ratio is set to a number randomly generated in the range of [0, 1]. The algorithm randomly selects a BS in $V$ to place the $i$-th VNF in $F^t$ until all VNFs in $F^t$ are placed.
	
\end{itemize}

\begin{figure}[!t]
	\centering
	\includegraphics[width=3.2in]{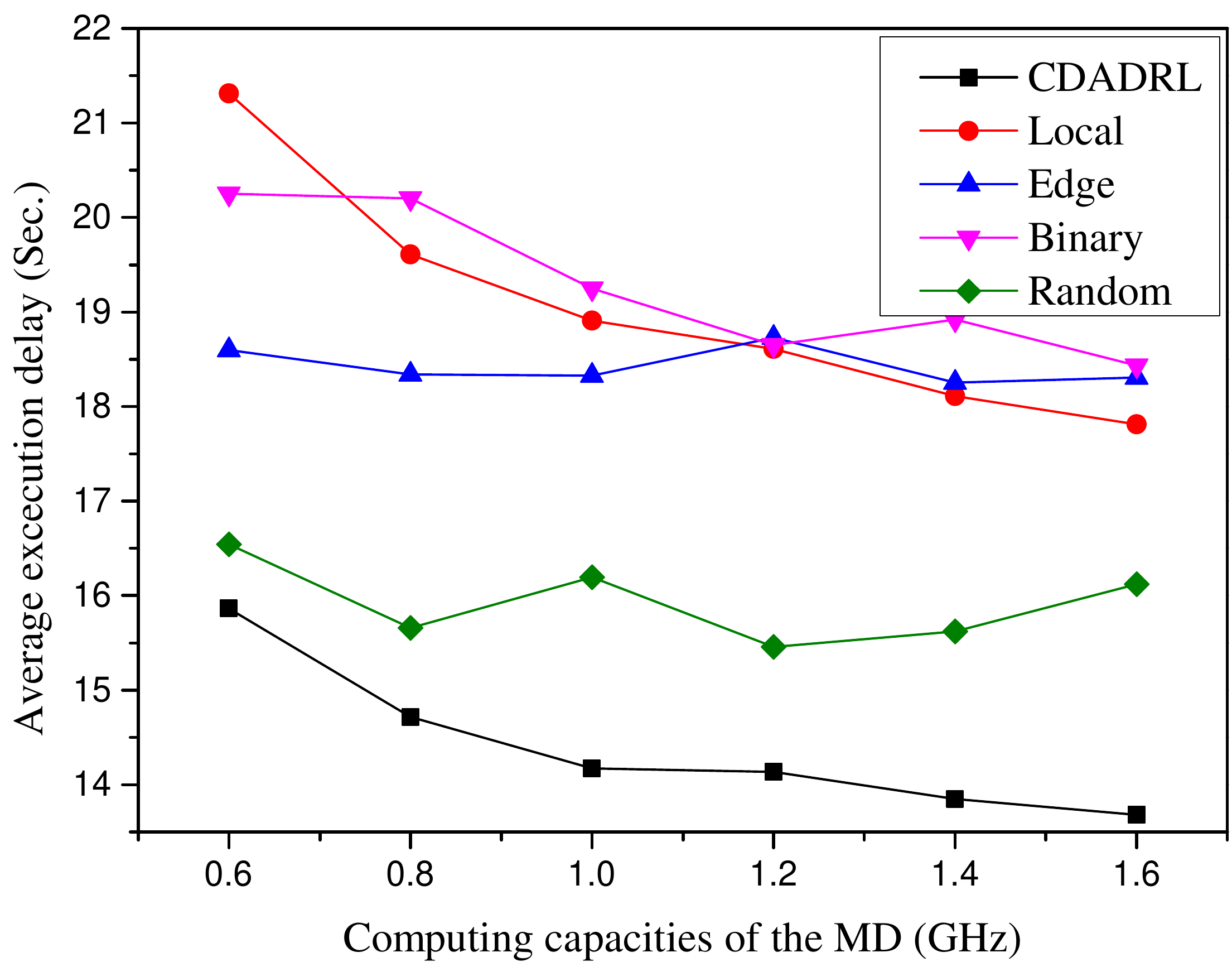}
	\caption{Results of average execution delay.}
	\label{fig-AED}
\end{figure}

\begin{figure}[!t]
	\centering
	\includegraphics[width=3.2in]{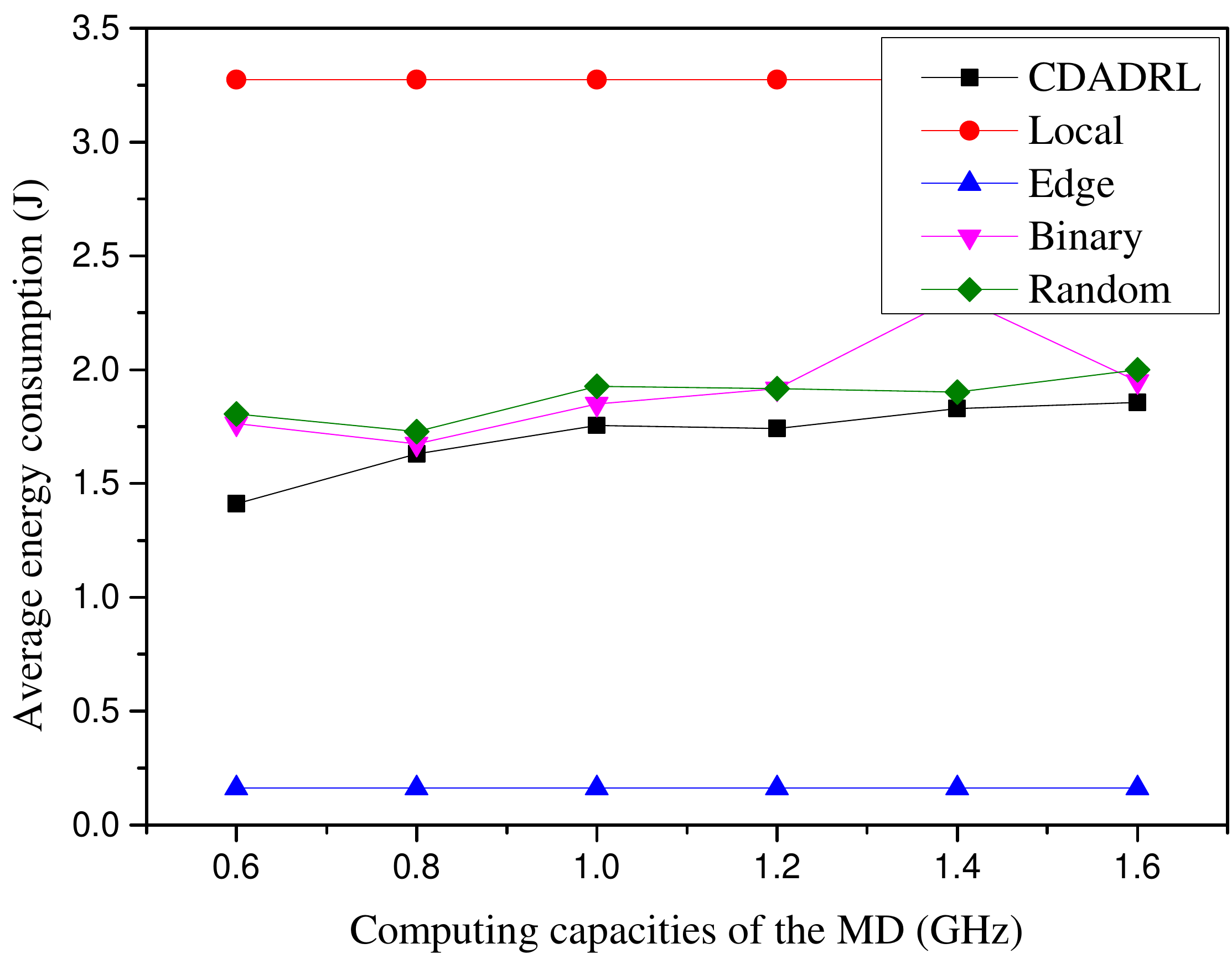}
	\caption{Results of average energy consumption.}
	\label{fig-AEC}
\end{figure}

\begin{figure}[!t]
	\centering
	\includegraphics[width=3.2in]{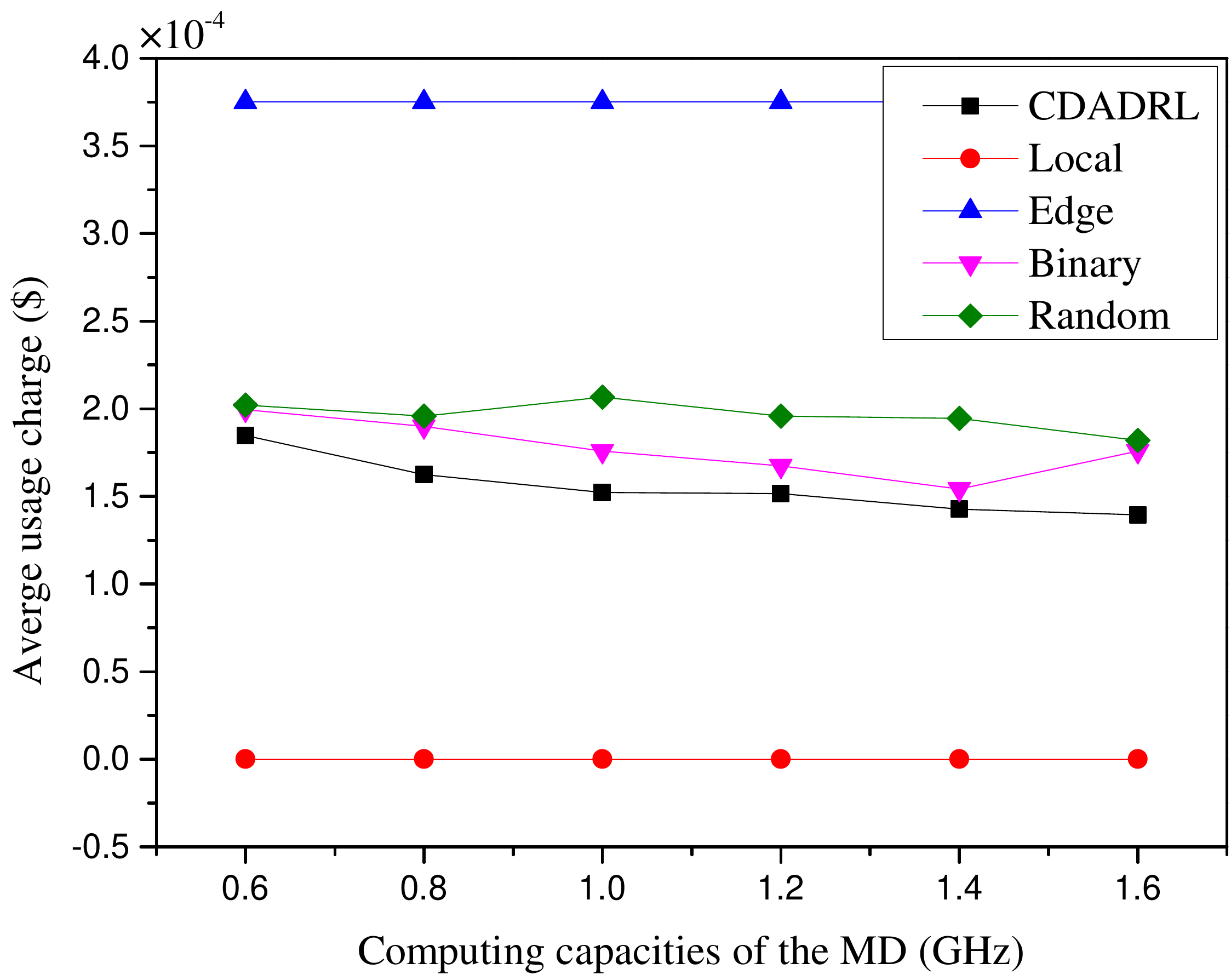}
	\caption{Results of average usage charge.}
	\label{fig-AUC}
\end{figure}

Figs.~\ref{fig-AED}-\ref{fig-AUC} show the offloading performance of algorithms with different offloading schemes, in terms of execution delay, energy consumption of the MD, and usage charge for edge computing. 
Fig.~\ref{fig-AED} depicts the results of average execution delay (AED) of the five algorithms with different computing capacities of the MD. It is observed that in most cases the AED value decreases as the MD's computing capacity increases. It is easy to understand that an MD with higher computing capacity offers a faster task processing speed, which lowers the execution delay. Meanwhile, the group number $\psi^t$ of $F^t$ is reduced to 1 when the MD's computing capacity is sufficient, decreasing the local computing delay $DL^t$. Obviously, \textbf{CDADRL} achieves the best AED value in each $cp_{MD}$ case. This is because \textbf{CDADRL} makes better decisions on partial offloading and VNF placement simultaneously, compared with \textbf{Local}, \textbf{Edge}, \textbf{Binary}, and \textbf{Random}.

The results of average energy consumption (AEC) are illustrated in Fig.~\ref{fig-AEC}. Contrary to the overall trend of AED vs. $cp_{MD}$, the AEC value increases with the MD's computing capacity growing up. As shown in Fig.~\ref{fig-AEC}, \textbf{Edge} and \textbf{Local} obtain the best and worst AEC values, respectively. In \textbf{Edge}, whenever a task request arrives, it is offloaded to the MEC servers for processing. In this case, the energy consumption incurred on the MD only involves transmitting the task data to the MEC servers and receiving the processed task from them. As for \textbf{Local}, any incoming task will be processed on the MD, leading to the highest energy consumption. \textbf{CDADRL} outperforms \textbf{Local}, \textbf{Binary}, and \textbf{Random} in all test scenarios.

Fig.~\ref{fig-AUC} shows the results of average usage charge (AUC). It is observed that \textbf{Edge} and \textbf{Local} obtain the worst and best results, respectively. \textbf{Local} always costs \$0 for edge computing in all test scenarios, as expected. \textbf{Edge} offloads all required tasks to the MEC servers, occupying more computing resources than the other algorithms. This is the rationale behind the highest AUC values. \textbf{CDADRL} performs better than \textbf{Edge}, \textbf{Binary}, and \textbf{Random} as it achieves smaller AUC values in all test scenarios.

\begin{figure}[!t]
	\centering
	\includegraphics[width=3.2in]{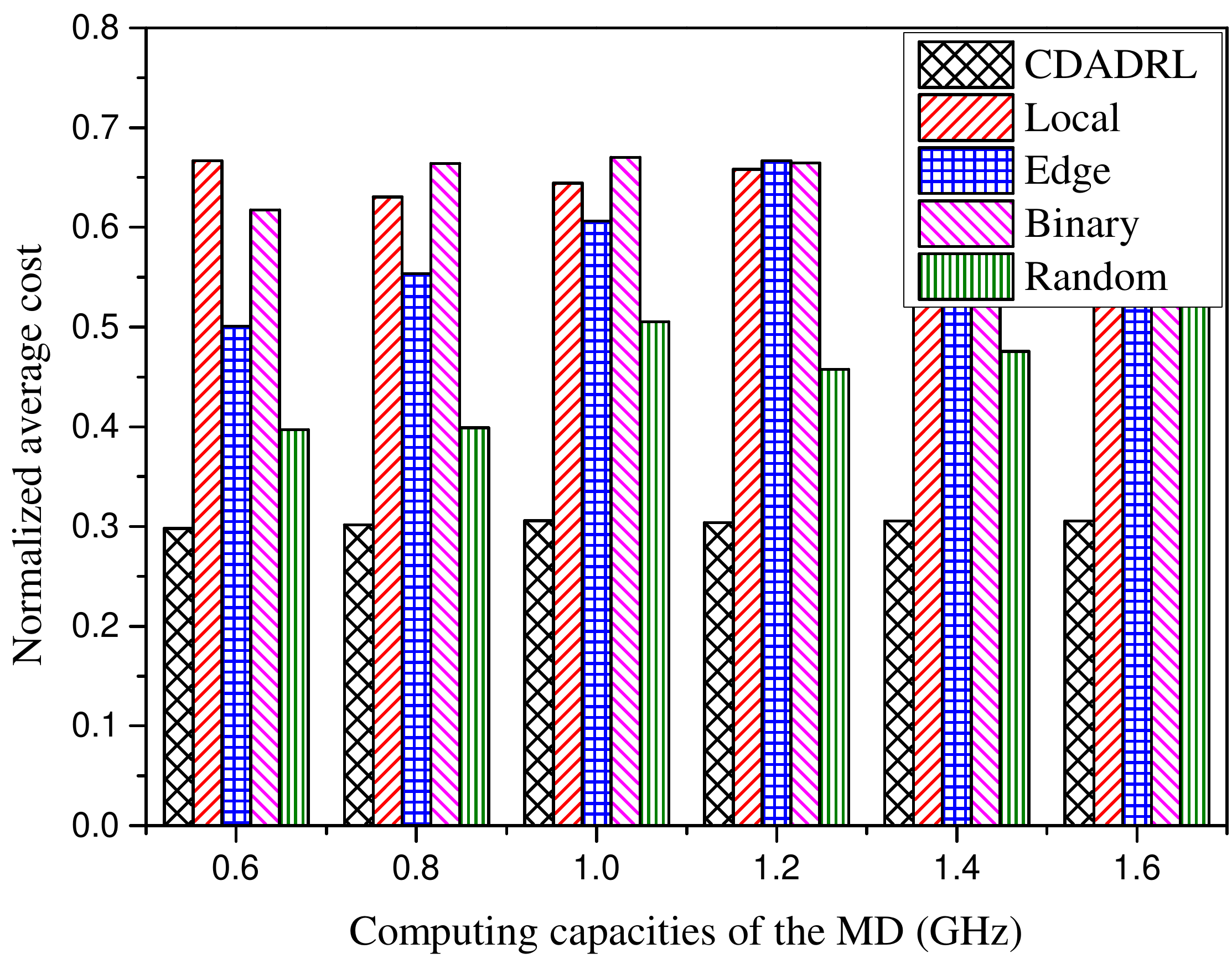}
	\caption{Results of normalized average cost.}
	\label{fig-COST}
\end{figure}

The normalized average cost (NAC) obtained by the five algorithms are shown in Fig.~\ref{fig-COST}. One can see clearly that \textbf{CDADRL} is the best algorithm. In addition, Friedman test \cite{ross2020introduction} is utilized for algorithm performance comparison. Based on the AED, AEC, AUC, and NAC, the average rankings of five algorithms in all test scenarios are shown in Table~\ref{tab:Ranking}. Obviously, the proposed \textbf{CDADRL} achieves the best offloading performance among the five algorithms.

\begin{table*}[!t]
	\caption{Ranking of five algorithms with different offloading schemes. \label{tab:Ranking}}
	\centering
	\begin{tabular}{lllllllll}
		\toprule
		\multirow{2}[4]{*}{Algorithm} & \multicolumn{2}{c}{AED} & \multicolumn{2}{c}{AEC} & \multicolumn{2}{c}{AUC} & \multicolumn{2}{c}{NAC} \\
		\cmidrule{2-9}          & Average rank & Position & Average rank & Position & Average rank & Position & Average rank & Position \\
		\midrule
		CDADRL & 1.0000 & 1     & 2.1667 & 2     & 2.0000 & 2     & 1.0000 & 1 \\
		Local & 4.1667 & 5     & 5.0000 & 5     & 1.0000 & 1     & 3.6667 & 3 \\
		Edge  & 3.8333 & 3     & 1.0000 & 1     & 5.0000 & 5     & 3.6667 & 3 \\
		Binary & 4.0000 & 4     & 2.8333 & 3     & 3.3333 & 3     & 4.6667 & 5 \\
		Random & 2.0000 & 2     & 4.0000 & 4     & 3.6667 & 4     & 2.0000 & 2 \\
		\bottomrule
	\end{tabular}
\end{table*}

\section{Conclusion and Future Work}
\label{Conclusion}
This paper formulates a partial offloading and SFC mapping joint optimization (POSMJO) problem in the context of NFV-enabled MEC. This problem consists of two closely related decision-making steps, namely task partition and VNF placement. The proposed cooperative dual-agent DRL (CDADRL) algorithm employs two agents to realize the two steps above. Effective collaboration between these agents is achieved through information exchange between the MD and EI environments. The TD3-based agent is able to provide high-quality task partition while the dueling DDQN-based agent offers appropriate placement for service function chain requests. Simulation results demonstrates the superiority of the proposed CDADRL over three combinations of state-of-the-art DRL algorithms with respect to cumulative reward and average episodic reward. Besides, CDADRL significantly outperforms a number of baseline algorithms with various offloading schemes in terms of execution delay, energy consumption, and usage charge.

In the future, we plan to adapt the proposed CDADRL algorithm to large-scale networks with multiple MDs, where each MD requests the partial offloading and SFC mapping services. In this paper, the execution delay, energy consumption, and usage charge are aggregated into one objective function to minimize. We assume the weights per objective are unchanged during learning and execution. However, in some cases, the weights are not known beforehand and may change over time. Single-objective RLs, e.g., DDPG, DQN, and CDADRL, are not suitable for these scenarios. Thus, we hope to further extend our algorithm to a multi-objective RL method.

\bibliographystyle{IEEEtran}
\bibliography{myref}

\vfill

\end{document}